\documentclass[10pt,twocolumn,letterpaper]{article}

\usepackage{cvpr}              

\usepackage[dvipsnames]{xcolor}

\usepackage{amsmath,graphicx}
\usepackage{amssymb}
\usepackage{svg} 
\usepackage{tabto}
\usepackage{color} 
\usepackage{multirow, tabularx}

\definecolor{cvprblue}{rgb}{0.21,0.49,0.74}
\usepackage[pagebackref,breaklinks,colorlinks,citecolor=cvprblue]{hyperref}

\def\paperID{13439} 
\def\confName{CVPR}
\def\confYear{2024}

\title{
Misalignment-Robust Frequency Distribution Loss for Image Transformation 
}

\author{
Zhangkai Ni$^1$, 
Juncheng Wu$^{1\dag}$, 
Zian Wang$^{1\dag}$, 
Wenhan Yang$^{2\ast}$, 
Hanli Wang$^{1\ast}$, 
Lin Ma$^3$  \\
$^1$Tongji University, 
$^2$Peng Cheng Laboratory, 
$^3$Meituan
}

\begin{document}

\twocolumn[{%
   \renewcommand\twocolumn[1][]{#1}%
   \maketitle
   
   \begin{center}
    \centering
    \includegraphics[width=0.244\textwidth]{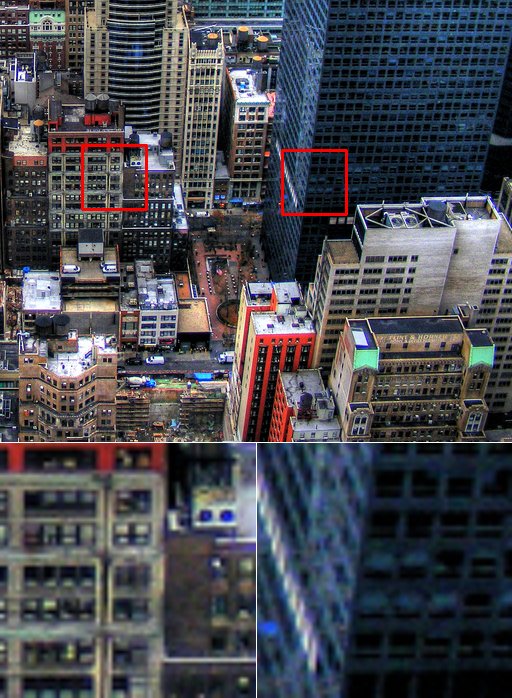}
    \includegraphics[width=0.244\textwidth]{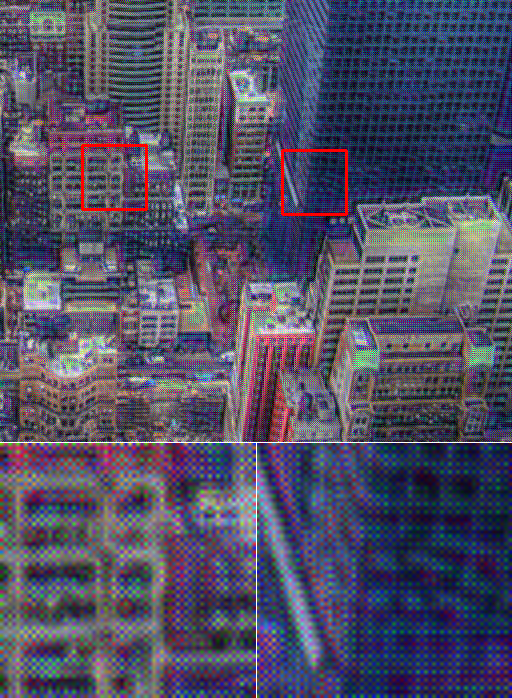}
    \includegraphics[width=0.244\textwidth]{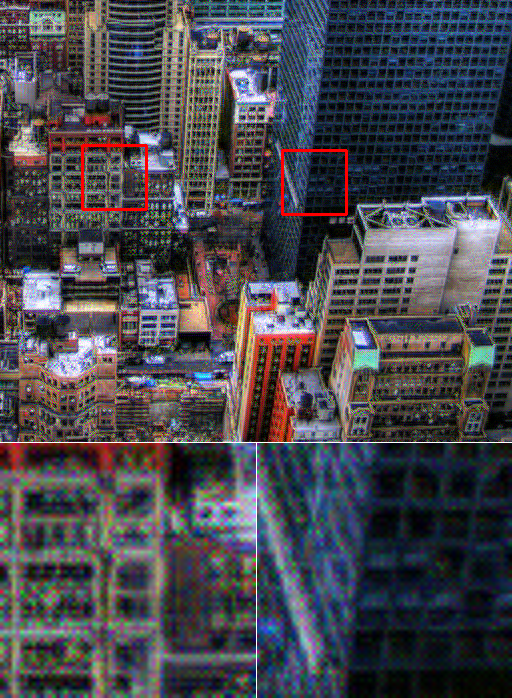}
    \includegraphics[width=0.244\textwidth]{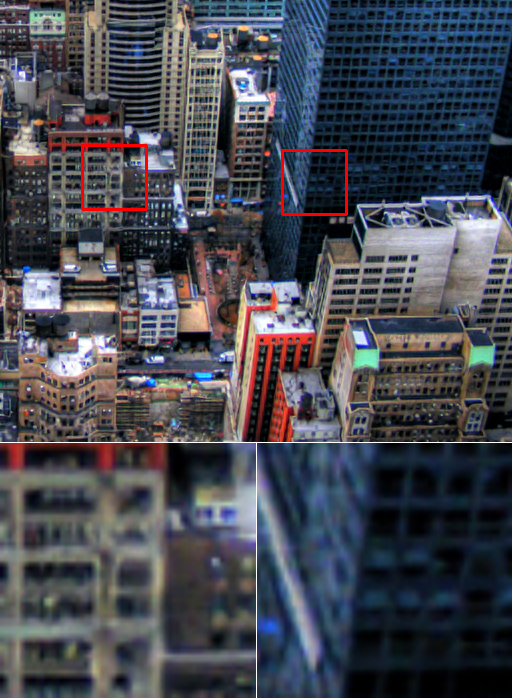}
    \rotatebox{0}{~~~~~~~~GT~~~~~~~~~~~~~~~~~~~~~~~~~~~~~~~~~~~~~~~~LPIPS~~~~~~~~~~~~~~~~~~~~~~~~~~~~~~~~~~~~~~~~PDL~~~~~~~~~~~~~~~~~~~~~~~~~~~~~~~~~~~~FDL (Ours)}

    \captionof{figure}{
    Qualitative results trained on our synthetic DIV2K dataset with strong misalignments. Compared with LPIPS~\cite{zhangUnreasonableEffectivenessDeep2018}, and PDL~\cite{delbracioProjectedDistributionLoss2021}, 
    the proposed method FDL yields clearer results with fewer artifacts. Zoom in to observe details.
 }
    \label{fig:headline}
   \end{center}
}]

\maketitle

\let\thefootnote\relax\footnotetext{\noindent$^\dag$Equal contribution. $^\ast$Corresponding author.}

\begin{abstract}

This paper aims to address a common challenge in deep learning-based image transformation methods, such as image enhancement and super-resolution, which heavily rely on precisely aligned paired datasets with pixel-level alignments.
However, creating precisely aligned paired images presents significant challenges and hinders the advancement of methods trained on such data.
To overcome this challenge, this paper introduces a novel and simple Frequency Distribution Loss (FDL) for computing distribution distance within the frequency domain.
Specifically, we transform image features into the frequency domain using Discrete Fourier Transformation (DFT).
Subsequently, frequency components (amplitude and phase) are processed separately to form the FDL loss function.
Our method is empirically proven effective as a training constraint due to the thoughtful utilization of global information in the frequency domain.
Extensive experimental evaluations, focusing on image enhancement and super-resolution tasks, demonstrate that FDL outperforms existing misalignment-robust loss functions.
Furthermore, we explore the potential of our FDL for image style transfer that relies solely on completely misaligned data.
Our code is available at: \url{https://github.com/eezkni/FDL}

\end{abstract}

\section{Introduction}
\label{sec:intro}

Image transformation refers to the process of changing the visual appearance or characteristics of images to achieve specific goals or effects. 
Various studies \cite{dongLearningDeepConvolutional2014,gharbi2017deep, ni2020towards,zhangImageSuperResolutionUsing2018,mei2021image} have showcased impressive results by integrating deep neural networks into image transformation tasks. 
For example, single image super-resolution (SISR) aims to enhance the spatial properties of images, while image enhancement strives to improve the quality, visibility, interpretability, etc.
However, a key limitation of these methods is the implicit assumption of pixel-aligned training data, thereby restricting their scope of applicability. 
This assumption is problematic as not all image transformation tasks can access perfectly aligned training data, particularly those involving natural distortions.
Besides, one prominent example is style transfer, a task that involves optimizing the style distance between images that lack content-related associations.
The misalignment of content in training data significantly challenges the effectiveness of these methods.

To mitigate this challenge, a considerable body of research has emerged focusing on the development of loss functions to improve the performance of image transformation models, including~\textit{element-wise} loss~\cite{johnson2016perceptual, zhangUnreasonableEffectivenessDeep2018, mechrezContextualLossImage2018} and~\textit{distribution-based} loss~\cite{delbracioProjectedDistributionLoss2021,heitz2021sliced,patchSWD}.
Element-wise loss functions are often ill-suited for geometric misaligned training data since even imperceptible misalignment can trigger significant responses to these losses.
Mechrez \etal~\cite{mechrezContextualLossImage2018} tackled the problem of misaligned training data through a patch-matching strategy, resulting in positive outcomes, particularly in misaligned tasks such as semantic style transfer. 
However, it often introduces artifacts under certain conditions since the spatial structure is ignored~\cite{zhangZoomLearnLearn2019,delbracioProjectedDistributionLoss2021}.
Distribution-based loss functions show promise in mitigating the interference of misaligned data, achieving better perceptual quality, and producing more realistic predictions in misaligned scenarios~\cite{delbracioProjectedDistributionLoss2021}.
However, these distribution-based loss functions often ignore spatial location, leading to potential structural error in predicted results.

In this paper, we present a comprehensive analysis of the distribution distance and propose solutions to address its limitations in accurately capturing the structural integrity of images (as shown in Section.~\ref{subsec:freq_structure}).
The previous work has demonstrated that the frequency domain contains more global information~\cite{jiangFocalFrequencyLoss2021,FDIT,yang2020fda,zhouSpatialFrequencyDomainInformation2022,huangDeepFourierBasedExposure2022}.
Our analysis shows that computing distribution distances in the frequency domain, as opposed to the spatial domain, can effectively leverage global information.
Consequently, when used as a training constraint, this approach can reduce structural errors in the predicted results.
Furthermore, frequency components of images possess various physical meanings~\cite{phase1,phase2}. 
In this work, we observe that the frequency components of image features encompass multiple characteristics within the image.
Therefore, integrating information from various frequency components in the loss function can better ensure the overall quality of the predicted images.

As a result, we propose a novel Frequency Distribution Loss (FDL) for image transformation models trained with misaligned data, opening up new avenues for addressing the broad issue of misalignment in image transformation tasks.
Specifically, we employ a pre-trained feature extractor to transform images (\textit{i.e.,} predicted image and target image) into the feature space.
Subsequently, two frequency components (amplitude and phase) are obtained individually from the predicted image features and the target image features using the Discrete Fourier Transform (DFT).
%
Finally, we employ Sliced Wasserstein Distance (SWD) to measure the distribution distance between the frequency components of predicted and target image features, respectively.
We conduct extensive experiments across various image transformation tasks, including single-image super-resolution, image enhancement, and style transfer, to demonstrate the effectiveness of FDL. FDL consistently achieves state-of-the-art performance in all evaluated scenarios, showcasing remarkable robustness to both models and tasks.
As illustrated in Figure~\ref{fig:headline}, FDL adeptly assesses the differences among essential information for SISR, even in the presence of strong geometric misalignment, ensuring the comprehensive quality of the predicted image.

\section{Related Work}
\label{sec:related}

\paragraph{Element-wise Losses for Image Transformation.} These loss functions calculate differences between pixels or features of images using an element-wise approach (\textit{e.g.}, L1 or L2 norm, Cosine distance). 
In many image transformation tasks, this type of loss proves effective in reducing distortion in the predicted image and ensuring its detail fidelity~\cite{blauPerceptionDistortionTradeoff2018, zhangUnreasonableEffectivenessDeep2018, johnson2016perceptual, zhao2016loss}.
However, when dealing with misaligned training data, even imperceptible small geometric variations can result in significant responses in such loss functions.
This lack of robustness to misalignment can lead to regression to the mean phenomenon in misaligned situations~\cite{delbracioProjectedDistributionLoss2021}, which makes it challenging to ensure the quality of predicted images.
To address this issue, several efforts have been made to enhance the robustness of feature extractors to geometric misalignment, including techniques such as anti-aliasing and max-pooling~\cite{ghildyalShiftTolerantPerceptualSimilarity2022,zhangMakingConvolutionalNetworks2019,kettunenELPIPSRobustPerceptual2019}. 
These improvements have shown promise, particularly in image quality assessment tasks. 
However, these modifications to models can lead to information loss, which poses challenges when employing them as loss functions in image transformation tasks. 
This limitation hinders the assurance of maintaining the quality of predicted images.
Mechrez~\etal. proposed Contextual Loss (CTX) by treating image features as a collection of patches and assessing the similarity between two input images through the calculation of element-wise distances between each feature patch and its nearest neighbor~\cite{mechrezContextualLossImage2018}.
The CTX loss brings a simple solution to misaligned data, however, since CTX cannot effectively utilize the global structural information of the image, artifacts may still appear in the predicted image~\cite{zhangZoomLearnLearn2019}.

\paragraph{Distribution-based Losses for Image Transformation.} 
These loss functions leverage distribution distances or disparities, such as Wasserstein Distance (WD)~\cite{patchSWD,nguyenDISTRIBUTIONALSLICEDWASSERSTEINAPPLICATIONS2021} or Kullback–Leibler divergence (KLD)~\cite{creswell2018generative}, to quantify the differences between image datasets or instances.
Initially used in image generation tasks, these loss functions have gained widespread application in transformation tasks. 
Empirical evidence has shown a strong correlation between these metrics and the perceptual quality of images~\cite{blauPerceptionDistortionTradeoff2018}.
GAN loss can be applied to completely misaligned data, it is often susceptible to introducing artifacts in the predicted images because GAN optimizes the distance between two image set-level distribution~\cite{ni2022cycle,ni2020towards}. 
In contrast, Elnekave \etal\cite{patchSWD} directly addresses the preservation of quality in the predicted image by matching the patch distribution of images.
Similarly, PDL~\cite{delbracioProjectedDistributionLoss2021} calculates the distance between the distribution of image features. 
These metrics based on spatial domain distribution distance at the image level show robustness to misalignment and can better ensure the quality of images.
However, these distribution-based measures only focus on distribution and ignore spatial location information.
Therefore, when using distribution distance as the loss function, it is hard to preserve structural accuracy in the predicted results.

\section{Methodology}
\label{sec:method}

\subsection{Overview}
\begin{figure}[t]
\centering
\includegraphics[width=1.0
\linewidth]{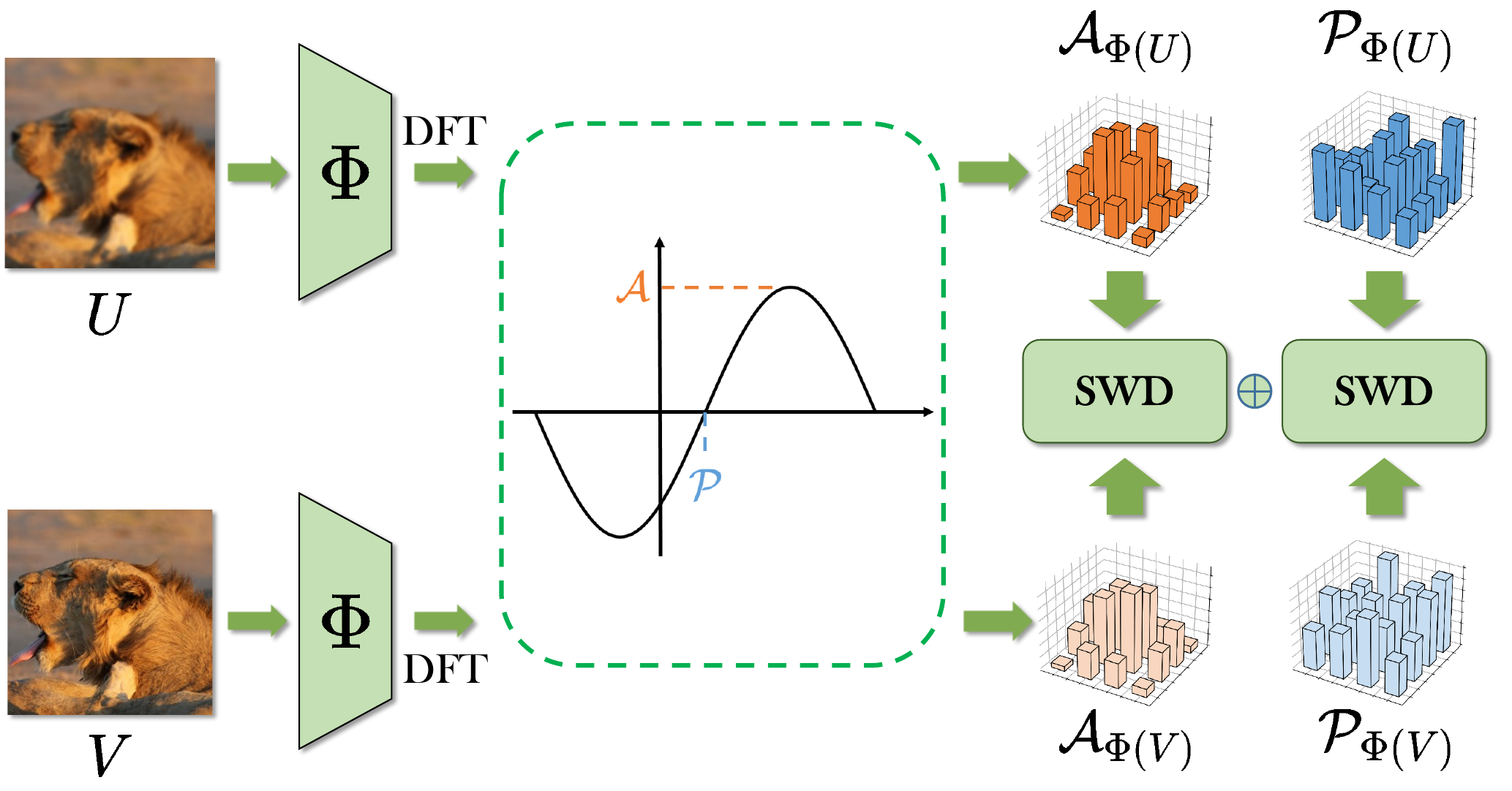}
\caption{
An overview of the proposed Frequency Distribution Loss (FDL). 
A shared feature extractor network $\Phi$ is utilized to project images into perceptual feature space.
Subsequently, the amplitude and phase of image features are obtained by Discrete Fourier Transform (DFT).
Then, the Sliced Wasserstein Distance (SWD)~\cite{patchSWD}, as an approximation of WD, is performed separately for amplitude and phase, and the results are linearly combined.
}
\label{fig:framework}
\end{figure}

We aim to design a loss function tailored for image transformation tasks, capable of measuring the similarity between misaligned images to ensure the overall quality.
In Section.~\ref{subsec:freq_structure}, we conduct a comprehensive analysis of the merits and challenges associated with distribution-based loss functions for image transformation tasks involving misaligned data.
We empirically demonstrate that computing distribution distances in the frequency domain can alleviate the challenge of disregarding positional information when calculating distribution distances in the spatial domain, thereby preserving the structural integrity of the predicted results.
In Section.~\ref{subsec:freq_decouple}, we explore the diverse information inherent in frequency components of image features.
Specifically, we demonstrate that two frequency components in the image feature space (amplitude and phase) are related to various characteristics of the image.
%
%
Therefore, integrating information in these frequency components is capable of ensuring the quality of the image in various aspects.
The overview of our proposed Frequency Distribution Loss (FDL) is shown in Figure~\ref{fig:framework}.

\subsection{Frequency Distribution Distance} 
\label{subsec:freq_structure}

Wasserstein Distance (WD) has been widely used to optimize neural networks by quantifying the dissimilarity between probability distribution.
It completely ignores spatial position information~\cite{kolouri2017optimal}, making it robust to geometric misalignment because it focuses on estimating the differences between the underlying distribution of the signals rather than their spatial alignment.
However, the disregard for spatial information may lead to the inability of WD to ensure the structural accuracy of the predicted results.
We argue that calculating WD in the spatial domain utilizes only the local information while utilizing global information can help address this issue.
Therefore, we anticipate that computing WD in the frequency domain better preserves the structural accuracy of predictions due to the richer global information presented in the frequency domain~\cite{FDIT,jiangFocalFrequencyLoss2021,zhouSpatialFrequencyDomainInformation2022}.

\begin{figure}[t]
    \centering 
\includegraphics[width=0.26\textwidth]{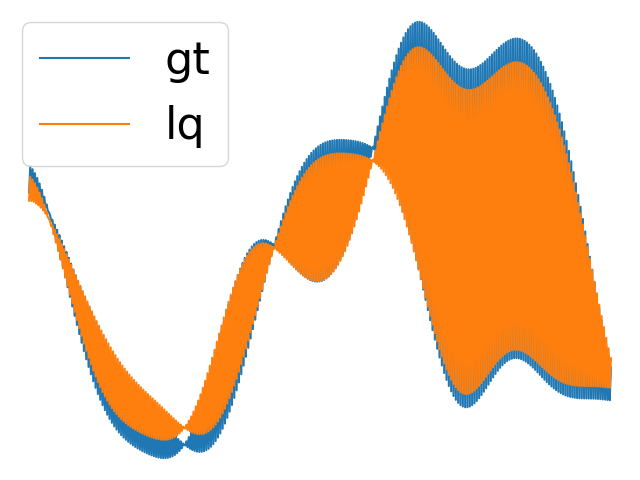}
\\
 \rotatebox{90}{\scriptsize{~~~~unaligned~~~~~~~~~~~~~~~~~~aligned}}
    \subfloat[$\mathcal{L}_{\text{MSE}}$]{
    \begin{minipage}[b]{0.14\textwidth}
        \includegraphics[width=1.0\textwidth]{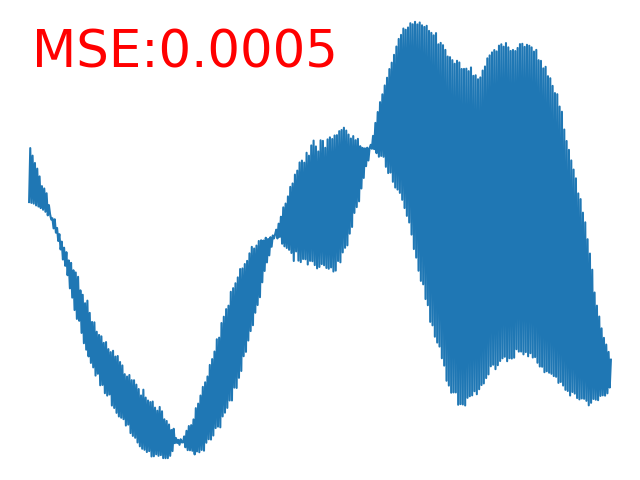}
        \\
        \includegraphics[width=1.0\textwidth]{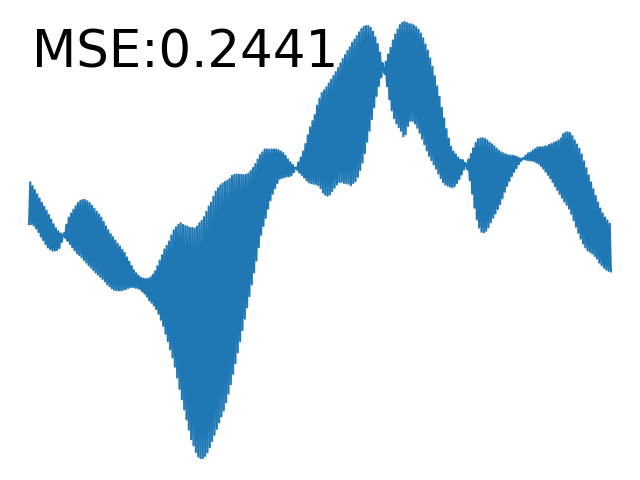}
    \end{minipage}
    }
    \subfloat[$\mathcal{L}_{\text{Spa}}$]{
    \begin{minipage}[b]{0.14\textwidth}
        \includegraphics[width=1.0\textwidth]{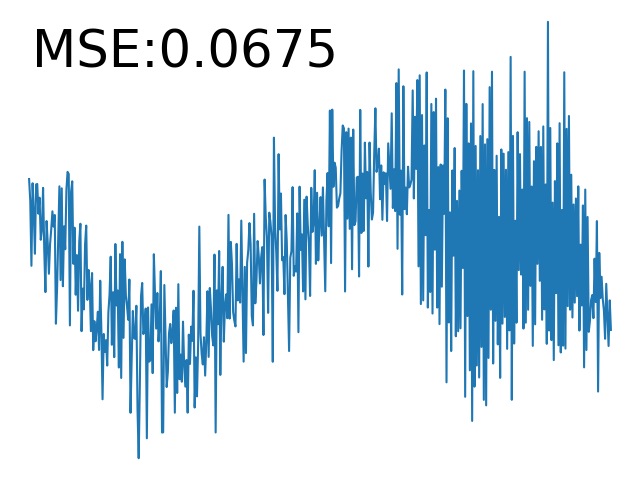}
        \\
        \includegraphics[width=1.0\textwidth]{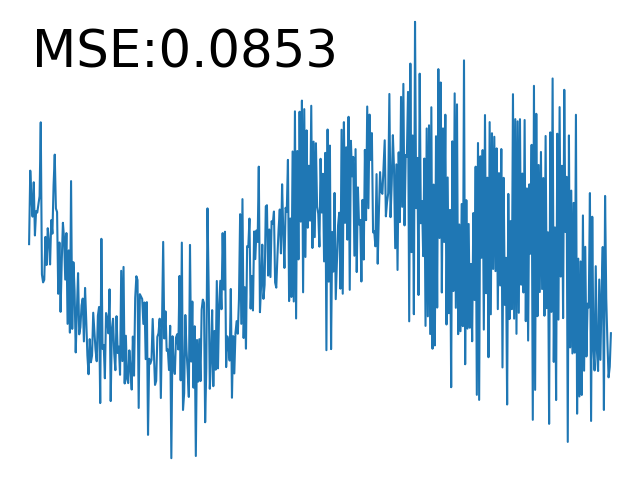}
    \end{minipage}
    }
    \subfloat[$\mathcal{L}_{\text{Freq}}$]{
    \begin{minipage}[b]{0.14\textwidth}
        \includegraphics[width=1.0\textwidth]{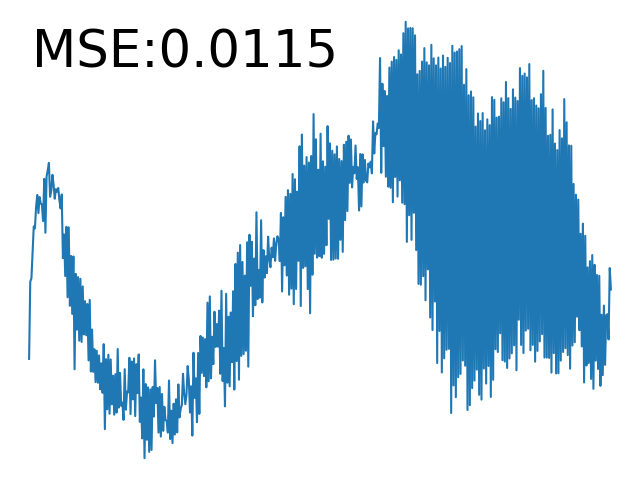}
        \\
        \includegraphics[width=1.0\textwidth]{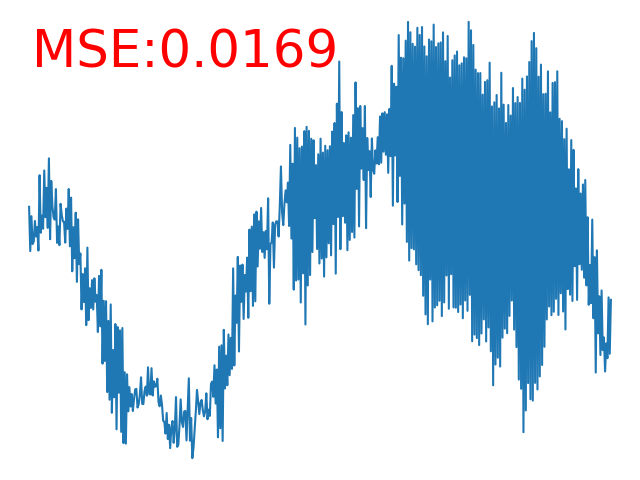}
    \end{minipage}
    }
    \hfill
	\caption{
    In the one-dimensional scenario, different loss functions are employed to train the same models with aligned and randomly misaligned training data, respectively. 
    $lq$ is the input test signal, and $gt$ is the corresponding ground truth.
    Each column represents the predicted results of models trained using different loss functions, with the MSE between the predicted result and ground truth.
	}
	\label{fig:one_dim}
\end{figure}

To validate this hypothesis, we conduct a straightforward toy experiment. 
Specifically, we generate a set of training data containing multiple pairs of one-dimensional input signals and their corresponding targets, where each pair of input and target signals has slightly different shapes (as shown in Figure~\ref{fig:one_dim}).
Additionally, we introduce random shifts to the target and input signals to induce misalignment within the training pairs.
We train a simple model $\text{M}(\cdot)$ to emulate the mapping from source to target, which can be formulated as:
\begin{equation}
\text{M}(x) = f(x) + x,
\label{eq:Linear}
\end{equation}
where $f(\cdot)$ represents a simple network, calculating the residual between the target and the input single $x$.

In the one-dimensional scenario, the WD between distribution has a closed-form solution. The loss function based on spatial domain WD can be formulated as:
\begin{equation}
    \mathcal{L}_{\text{Spa}}(\text{M}(x),y) = \text{WD}\left(\text{M}(x),y\right),
\label{eq:Linear_SWD}
\end{equation}
where $x$ and $y$ are the input and target signal, respectively, $\text{WD}(\cdot,\cdot)$ represents the one-dimensional Wasserstein Distance between the distribution of the signals.
To calculate WD in frequency domain, we initially utilize the Discrete Fourier Transform (DFT) to transform signals into the frequency domain, obtaining frequency components (amplitude and phase), which contain all the frequency domain information.
Next, the loss function based on frequency WD can be formulated as:
\begin{equation}
    \mathcal{L}_{\text{Freq}}(\text{M}(x),y) = \text{WD} \left(\mathcal{A}_{\text{M}(x)}
    , \mathcal{A}_{y}\right) + \text{WD}\left(\mathcal{P}_{\text{M}(x)}, \mathcal{P}_{y}\right),
\label{eq:Linear_FDL}
\end{equation}
where $\mathcal{A}_s = \left| \mathcal{F}\circ s \right|$ is the amplitude of the spectrum of signal $s$, and $\mathcal{P}_s = \angle \left( \mathcal{F}\circ s \right)$ is the phase, $\mathcal{F}$ denotes DFT.
%
%
We employ Mean Squared Error ($\mathcal{L}_{\text{MSE}}$), $\mathcal{L}_{\text{Spa}}$ and $\mathcal{L}_{\text{Freq}}$ as loss function to train the mode respectively.
And we conduct training with both aligned and misaligned training data.

The comparison results in Figure~\ref{fig:one_dim} show that directly using $\mathcal{L}_{\text{MSE}}$ enables the model to effectively learn the mapping from input to target when there is no misalignment in the training data. In contrast, when there is misalignment in the training data, models trained with $\mathcal{L}_{\text{MSE}}$ exhibit a significant decrease in prediction accuracy compared to perfectly aligned training data. Meanwhile, models trained with $\mathcal{L}_{\text{Spa}}$ and $\mathcal{L}_{\text{Freq}}$ as loss functions show less change in performance.
This indicates that both $\mathcal{L}_{\text{Spa}}$ and $\mathcal{L}_\text{Freq}$ exhibit shift robustness. 

However, $\mathcal{L}_{\text{Spa}}$ completely disregards spatial positional information, leading to the model output having a similar distribution to the target but not guaranteeing the structural accuracy of the prediction. Therefore, we turn to measure the WD of the frequency domain for better structural accuracy.

\subsection{Feature Frequency Components}
\label{subsec:freq_decouple}
In Section.~\ref{subsec:freq_structure}, we empirically demonstrate the benefit of calculating the distribution distance in frequency domain using amplitude and phase.
These frequency components of the image possess specific physical meanings~\cite{phase1,phase2}, thus we reckon that these frequency components of the image feature are likely to be associated with certain image characteristics.
In this section, we further investigate the information associated with amplitude and phase of image features. 
We provide empirical analysis through a simple experiment.
Specifically, we first extract features from two images denoted as $Q$ and $D$ using the encoder $\Phi(\cdot)$ based on VGG.
Next, we obtain the amplitude and phase of these two features through DFT respectively, and mix the amplitude of $\Phi(Q)$ and the phase of $\Phi(D)$.
Subsequently, we project the mixed amplitude and phase back into the feature domain through inverse DFT and adopt a decoder to transform the feature obtained from the mixed frequency components back into pixel space.
The pretext process can be expressed as:
\begin{equation}
res = \Phi^{-1}\left(\mathcal{F}^{-1}\circ \left( \mathcal{A}_{\Phi(Q)},\mathcal{P}_{\Phi(D)} \right) \right),
\label{eq:swap_toy}
\end{equation}
where $\Phi (\cdot)$ and $\Phi^{-1} (\cdot)$ are the encoder and decoder respectively. $\mathcal{F}^{-1}$ denotes the inverse DFT and $res$ is the generated images.

\begin{figure}[t]
    \centering 
    \subfloat[$Q$]{
    \begin{minipage}[b]{0.15\textwidth}
        \includegraphics[width=1.0\textwidth]{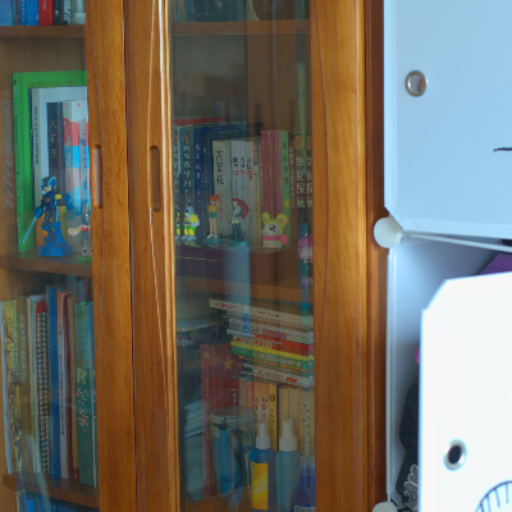}
        \\
        \includegraphics[width=1.0\textwidth]
        {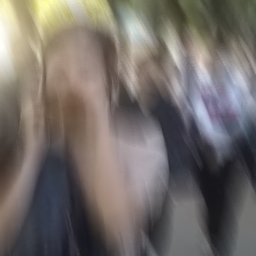}
    \end{minipage}
    }
    \subfloat[$D$]{
    \begin{minipage}[b]{0.15\textwidth}
        \includegraphics[width=1.0\textwidth]{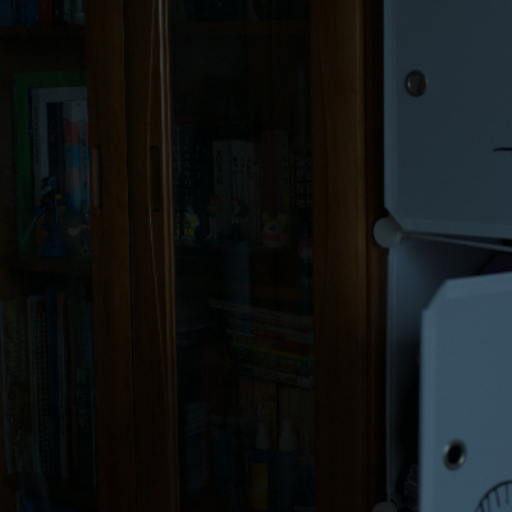}
        \\
        \includegraphics[width=1.0\textwidth]
        {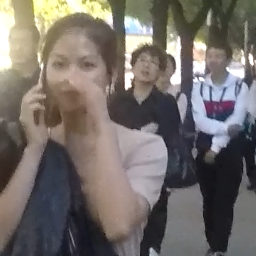}
    \end{minipage}
    }
    \subfloat[$res$]{
    \begin{minipage}[b]{0.15\textwidth}
        \includegraphics[width=1.0\textwidth]{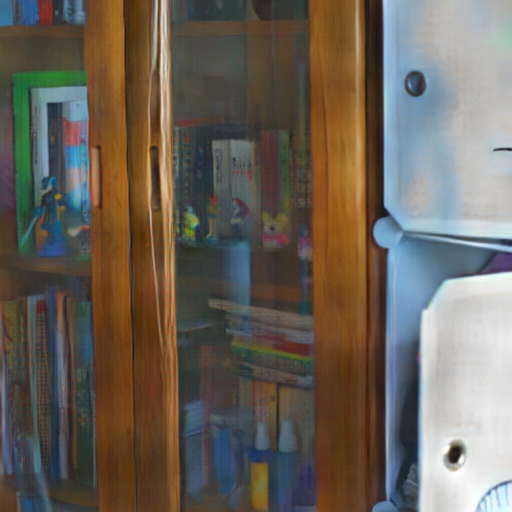}
        \\
        \includegraphics[width=1.0\textwidth]
        {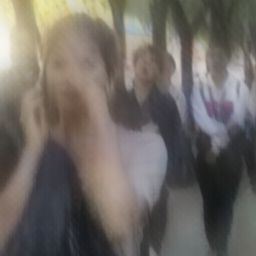}
    \end{minipage}
    }
    \hfill
	\caption{
		Result of frequency components mixing. An encoder ($\Phi$) extracts features from Q and D. We mix the frequency components using the amplitude of $\Phi(Q)$ and the phase of $\Phi(D)$. Then the feature with mixed-frequency component is decoded into the pixel domain.
	}
	\label{fig:swap_toy_example}
\end{figure}

Figure~\ref{fig:swap_toy_example} shows the experimental results. The images used in our experiment are sourced from the LOL~\cite{LOL} and HIDE~\cite{swapDeblur} datasets.
%
%
By comparing the generated images with image $Q$ and image $D$, we can discern the information associated with frequency components in the feature domain.
Observation shows that the resulting image's texture-related attributes like illumination and color resemble those of image $Q$, which provides amplitude. Meanwhile, the structural elements such as object shapes and edges exhibit high similarity between the result and image $D$.
We therefore argue that in the feature domain, the amplitude and phase component is associated with information related to various characteristics of the image. 
Therefore, we believe that incorporating the information in these frequency components in the loss function, allows for a comprehensive consideration of the various characteristics within the image, thereby enhancing the overall quality of the predicted results.

\subsection{Overall Loss Function}
\begin{figure}[t]
    \centering  
    \includegraphics[width=0.50 \textwidth]{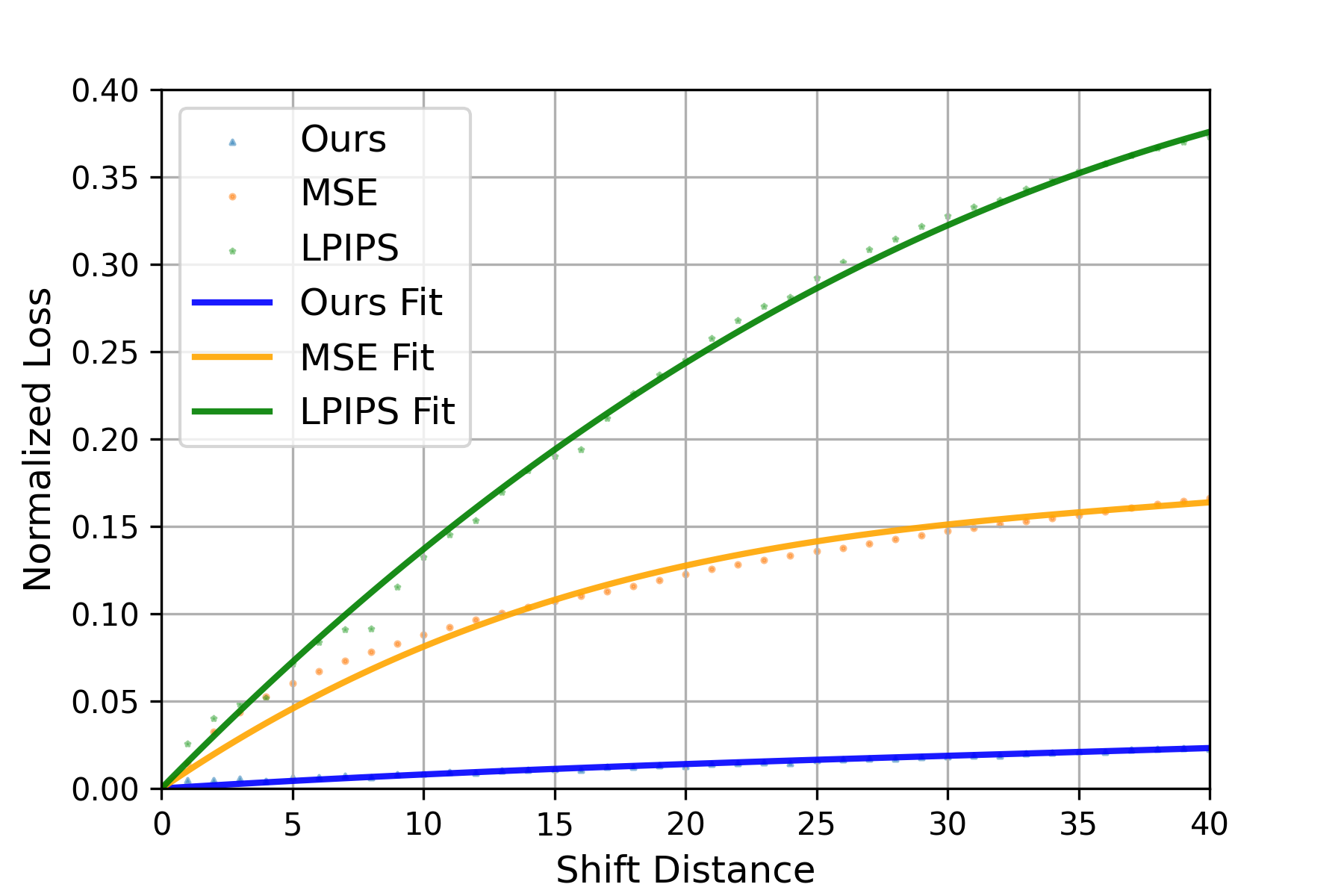}
	\caption{
		Shift response curves for different loss functions, including FDL, LPIPS, and Mean Square Error(MSE). We randomly shift the reference image for different pixels and calculate the discrepancy between the shifted and reference image using different metrics. The proposed FDL demonstrates strong shift robustness.
	}
	\label{fig:shift}
\end{figure}

We summarize the previous analysis and propose the Frequency Distribution Loss (FDL) between the predicted image and the target image, which can be formulated as:
\begin{equation}
\small
\mathcal{L}_{\text{FDL}}\left(U, V \right) = 
\text{SW}\left(\mathcal{A}_{\Phi(U)},\mathcal{A}_{\Phi(V)} \right) + \lambda\cdot \text{SW}\left(\mathcal{P}_{\Phi(U)},\mathcal{P}_{\Phi(V)} \right),
\label{eq:FDL}
\end{equation}
where $U$ and $V$ refer to predicted and target images, respectively.
$\Phi(\cdot)$ refers to an arbitrary feature extractor.
$\text{SW}(\cdot,\cdot)$ represents the Sliced Wasserstein Distance (SWD) between the distribution of two signals.
Due to the absence of a closed-form solution for WD in high-dimensional spaces, following the Elnekave \etal~\cite{patchSWD}, we employ the SWD as an approximation of WD.
As shown in Figure~\ref{fig:shift}, FDL exhibits strong shift invariance, making it suitable for various scenarios with geometric misalignment.

\section{Experiment}
\label{sec:exp}

To demonstrate the superiority and generality of our method, we adopt the proposed FDL into various image transformation tasks, including image enhancement, single image super-resolution, and style transfer. 
For each task, we employ multiple representative baseline models and ensure that each model is trained using only the proposed FDL or the compared loss functions.
Note that our focus is exclusively on scenarios where training data is misaligned.

\subsection{Experiment Settings}
\noindent \textbf{Baseline Models.} 
To comprehensively validate the robustness of our proposed FDL across different architectural models, we select various baseline models for each task: 1) NAFNet~\cite{chen2022simple} and SwinIR~\cite{liang2021swinir} for image enhancement; 2) NLSN~\cite{mei2021image}, NAFNet~\cite{chen2022simple} and SwinIR~\cite{liang2021swinir} for single image super-resolution (SISR); and 3) Gatys \etal~\cite{style_transfer} for style transfer. 
The NAFNet, NLSN and  Gatys \etal are convolutional neural networks (CNN) based models, while SwinIR is a Transformer~\cite{liu2021swin} based model.
These models have shown impressive results in corresponding tasks and have been recognized as representative models in recent years.

\noindent \textbf{Baseline Datasets.} 
For image enhancement, we choose the DPED~\cite{ignatovDSLRQualityPhotosMobile2017a} dataset for both training and testing. 
DPED exhibits significant geometric misalignment between the low-quality images and the high-quality image pairs because these image pairs are captured by different devices with the same scene.
Despite employing alignment algorithms and cropping the training images into smaller patches to minimize the impact of misalignment, visually noticeable misalignment still exists in this dataset. 
Additionally, DPED contains a substantial amount of real-world noise, posing challenges for both the models and the loss functions.
For SISR, we select the real-world SISR datasets for training and testing by combing the RealSR~\cite{cai2019toward} and City100~\cite{chen2019camera} datasets.
Furthermore, to examine the capability and generality of FDL in the presence of strong misalignment, a dataset with significant misalignment is synthesized based on the DIV2K dataset.
We randomly crop two images with noticeable geometric misalignment from the high-resolution image and downsample one of the cropped images to generate a low-resolution image. This low-resolution image is paired with the other cropped high-resolution image to train SISR models.
This is done to simulate irregular displacements that may occur in real-world scenarios.
\begin{figure*}[ht]
    \centering 
    \subfloat[Input]{
    \begin{minipage}[b]{0.19\textwidth}
        \includegraphics[width=1.0\textwidth]{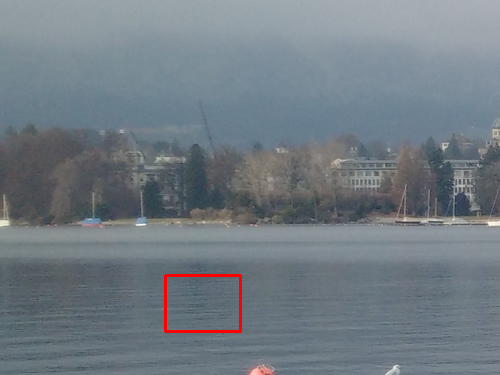}
        \\
        \includegraphics[width=1.0\textwidth]{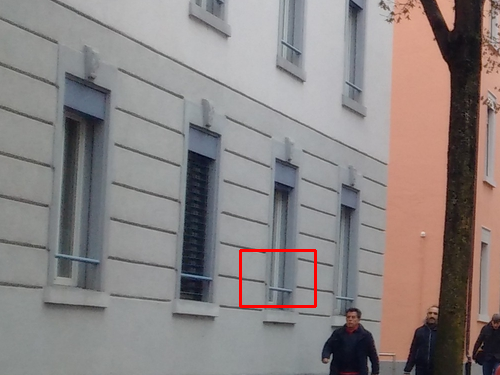}
        \\
        \includegraphics[width=1.0\textwidth]{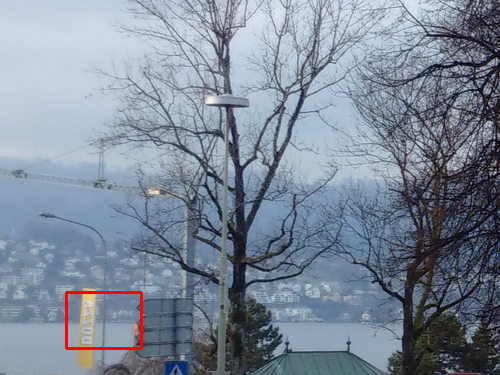}
    \end{minipage}
    }
    \subfloat[LPIPS]{
    \begin{minipage}[b]{0.19\textwidth}
        \includegraphics[width=1.0\textwidth]{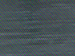}
        \\
        \includegraphics[width=1.0\textwidth]{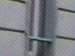}
        \\
        \includegraphics[width=1.0\textwidth]{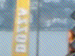}
    \end{minipage}
    }
    \subfloat[PDL]{
    \begin{minipage}[b]{0.19\textwidth}
        \includegraphics[width=1.0\textwidth]{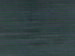}
        \\
        \includegraphics[width=1.0\textwidth]{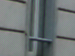}
        \\
        \includegraphics[width=1.0\textwidth]{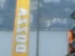}
    \end{minipage}
    }
    \subfloat[CTX]{
    \begin{minipage}[b]{0.19\textwidth}
        \includegraphics[width=1.0\textwidth]{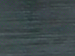}
        \\
        \includegraphics[width=1.0\textwidth]{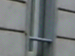}
        \\
        \includegraphics[width=1.0\textwidth]{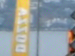}
    \end{minipage}
    }
    \subfloat[FDL (Ours)]{
    \begin{minipage}[b]{0.19\textwidth}
        \includegraphics[width=1.0\textwidth]{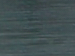}
        \\
        \includegraphics[width=1.0\textwidth]{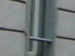}
        \\
        \includegraphics[width=1.0\textwidth]{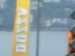}
    \end{minipage}
    }
    \vspace{-6pt}
    \hfill
	\caption{
		Qualitative results on DPED dataset~\cite{ignatovDSLRQualityPhotosMobile2017a} and NAFNet~\cite{chen2022simple} compared with LPIPS, PDL, CTX. The red area is cropped from different results and enlarged for visual convenient. Zoom in to observe details.
	}
	\label{fig:DPED_Comparison}
\end{figure*}

\noindent \textbf{Baseline Loss Functions.} 
We compare our proposed method with several state-of-the-art loss functions, including CTX~\cite{mechrezContextualLossImage2018}, PDL~\cite{delbracioProjectedDistributionLoss2021}, and LPIPS~\cite{zhangUnreasonableEffectivenessDeep2018}.
CTX and PDL are loss functions specifically designed for handling misaligned data.
LPIPS is a well-known and widely used perceptual loss in various image restoration tasks.
All the loss functions follow the official settings for fairness comparison.
Specifically, VGG19~\cite{Simonyan2014VeryDC} is used as the feature extractor for CTX and PDL, while AlexNet~\cite{krizhevsky2012imagenet} is utilized as the feature extractor for LPIPS.

\noindent \textbf{Evaluation Metrics.} 
We select PSNR, SSIM~\cite{wangImageQualityAssessment2004}, LPIPS~\cite{zhangUnreasonableEffectivenessDeep2018}, DISTS~\cite{dingImageQualityAssessment2020}, and FID~\cite{heusel2017gans} as the evaluation metrics for image enhancement and SISR. 
SSIM~\cite{wangImageQualityAssessment2004} and PSNR can assess the fidelity of image details, while DISTS~\cite{dingImageQualityAssessment2020}, FID~\cite{heusel2017gans}, and LPIPS~\cite{zhangUnreasonableEffectivenessDeep2018} reflect the perceptual quality of predicted images. 

\noindent \textbf{Implementation Details.}
In our work, we employ VGG19~\cite{Simonyan2014VeryDC} as the feature extractor and compute FDL on the $Relu\_1\_1$, $Relu\_2\_1$, $Relu\_3\_1$, $Relu\_4\_1$, and $Relu\_5\_1$ layers.
%
In different scenarios, we adjust the parameter $\lambda$ to modulate the weight assigned to different frequency components.
Specifically, for super-resolution and style transfer tasks, $\lambda=1$, while for the image enhancement task, $\lambda=0.01$.

\begin{table}[t]
  \centering
    \footnotesize
    \renewcommand\arraystretch{1.0}
    \setlength{\tabcolsep}{2.0pt}
    \begin{tabular}{c|c|ccccc}
    \hline
    \hline
    Model & Loss & PSNR$\uparrow$  & LPIPS$\downarrow$ & DISTS$\downarrow$ & SSIM$\uparrow$  & FID$\downarrow$ \\
    \hline
    \multirow{4}[2]{*}{NAFNet} & CTX & 22.256 & 0.126 & 0.148 & \textcolor[rgb]{ 0,  .439,  .753}{0.778} & \textcolor[rgb]{ 0,  .439,  .753}{74.553} \\
          & LPIPS(Alex) & 20.819 & 0.291 & 0.233 & 0.584 & 215.350 \\
          & PDL   & \textcolor[rgb]{ 0,  .439,  .753}{22.665} & \textcolor[rgb]{ 0,  .439,  .753}{0.117} & \textcolor[rgb]{ 0,  .439,  .753}{0.128} & 0.776 & 75.124 \\
          & \textbf{FDL (Ours)} & \textcolor[rgb]{ 1,  0,  0}{\textbf{23.048}} & \textcolor[rgb]{ 1,  0,  0}{\textbf{0.114}} & \textcolor[rgb]{ 1,  0,  0}{\textbf{0.121}} & \textcolor[rgb]{ 1,  0,  0}{\textbf{0.811}} & \textcolor[rgb]{ 1,  0,  0}{\textbf{37.501}} \\
    \hline
    \multirow{4}[2]{*}{SwinIR} & CTX & 20.800  & \textcolor[rgb]{ 0,  .439,  .753}{0.134} & \textcolor[rgb]{ 0,  .439,  .753}{0.152} & 0.734 & \textcolor[rgb]{ 0,  .439,  .753}{64.093} \\
          & LPIPS(Alex) & \textcolor[rgb]{ 1,  0,  0}{21.613} & 0.157 & 0.168 & \textcolor[rgb]{ 0,  .439,  .753}{0.759} & 127.310 \\
          & PDL   & 20.256 & 0.152 & 0.167 & 0.701 & 107.726 \\
          & \textbf{FDL (Ours)} & \textcolor[rgb]{ 0,  .439,  .753}{\textbf{21.488}} & \textcolor[rgb]{ 1,  0,  0}{\textbf{0.128}} & \textcolor[rgb]{ 1,  0,  0}{\textbf{0.136}} & \textcolor[rgb]{ 1,  0,  0}{\textbf{0.786}} & \textcolor[rgb]{ 1,  0,  0}{\textbf{29.877}} \\
    \hline
    \hline
    \end{tabular}%
    \caption{
    Quantitative comparison of image enhancement on the DPED dataset~\cite{ignatovDSLRQualityPhotosMobile2017a}.
    The best and second best results are marked in \textcolor[rgb]{ 1,  0,  0}{red} and \textcolor[rgb]{ 0,  .439,  .753}{blue}, respectively.
    }
  \label{tab:DPED}%
\end{table}%

\subsection{Image Enhancement}
The quantitative comparison results presented in Table~\ref{tab:DPED} demonstrate the superiority of our proposed FDL over all compared loss functions across various evaluation criteria. 
This indicates in the presence of significant geometric misalignment in the dataset, our loss function not only preserves fine details in the images but also achieves superior perceptual quality compared to existing misaligned loss functions.
Thus, the proposed FDL achieves a better perceptual distortion tradeoff~\cite{blauPerceptionDistortionTradeoff2018}.
These advantages can be attributed to the integration of frequency domain information in our loss function.
The visual comparison in Figure~\ref{fig:DPED_Comparison} provides several insightful observations.
The element-wise loss functions like CTX and LPIPS struggle to accurately capture differences in structured detail information, resulting in noticeable artifacts. 
In contrast, distribution based loss functions such as the proposed FDL and PDL can significantly reduce artifacts.
The limitations of PDL in accurately measuring more global differences in structural information arise from its calculation of distribution distances in the spatial domain.
Our proposed FDL addresses this limitation by calculating distribution distance in the frequency domain, which helps it successfully achieve excellent results in the presence of strong geometric misalignment.

\begin{figure*}[ht]
    \centering 
    \subfloat[Input]{
    \begin{minipage}[b]{0.19\textwidth}
        \includegraphics[width=1.0\textwidth]{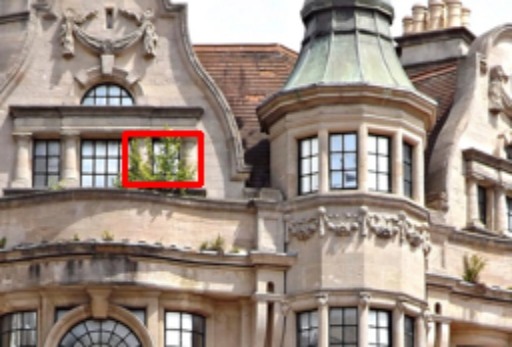}
        \\
        \includegraphics[width=1.0\textwidth]{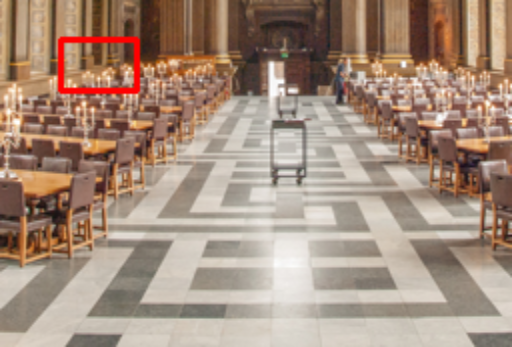}
        \\
        \includegraphics[width=1.0\textwidth]{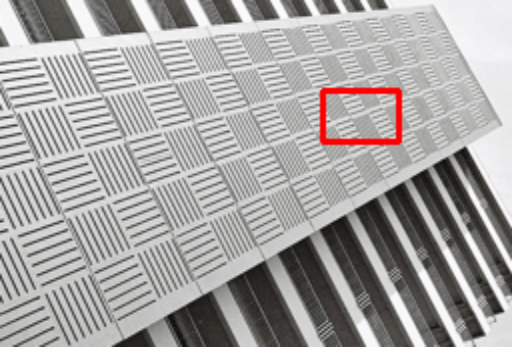}
    \end{minipage}
    }
    \subfloat[LPIPS]{
    \begin{minipage}[b]{0.19\textwidth}
        \includegraphics[width=1.0\textwidth]{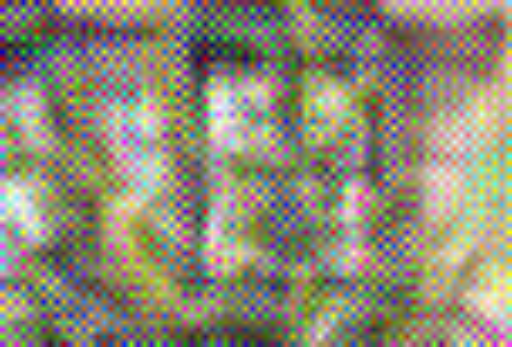}
        \\
        \includegraphics[width=1.0\textwidth]{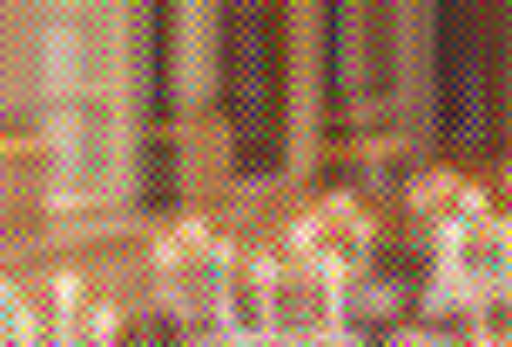}
        \\
        \includegraphics[width=1.0\textwidth]{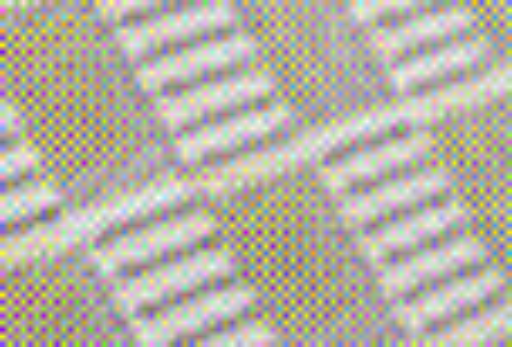}
    \end{minipage}
    }
    \subfloat[PDL]{
    \begin{minipage}[b]{0.19\textwidth}
        \includegraphics[width=1.0\textwidth]{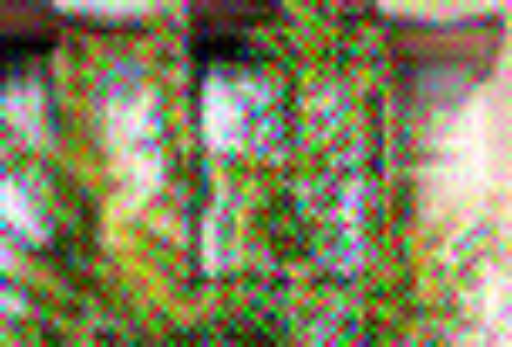}
        \\
        \includegraphics[width=1.0\textwidth]{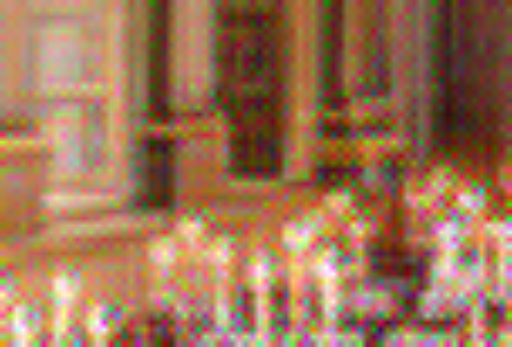}
        \\
        \includegraphics[width=1.0\textwidth]{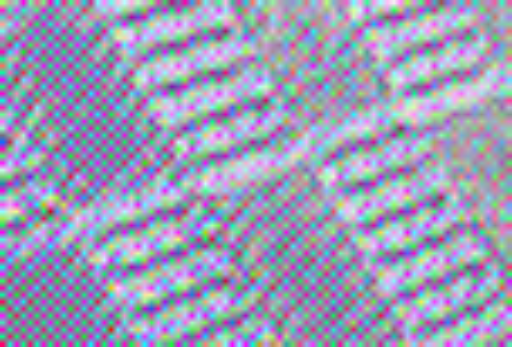}
    \end{minipage}
    }
    \subfloat[CTX]{
    \begin{minipage}[b]{0.19\textwidth}
        \includegraphics[width=1.0\textwidth]{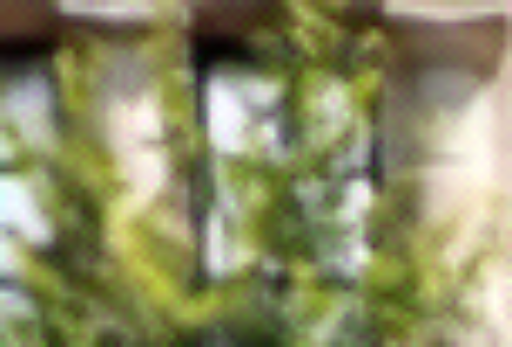}
        \\
        \includegraphics[width=1.0\textwidth]{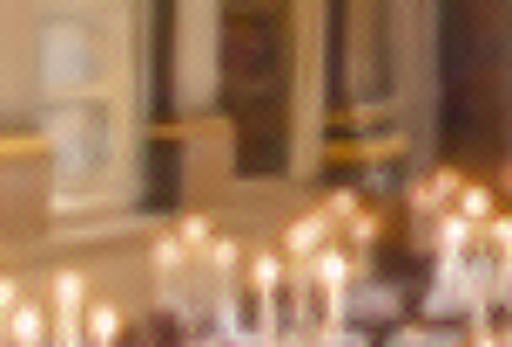}
        \\
        \includegraphics[width=1.0\textwidth]{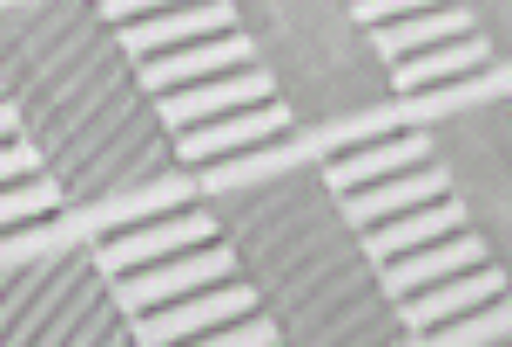}
    \end{minipage}
    }
    \subfloat[FDL (Ours)]{
    \begin{minipage}[b]{0.19\textwidth}
        \includegraphics[width=1.0\textwidth]{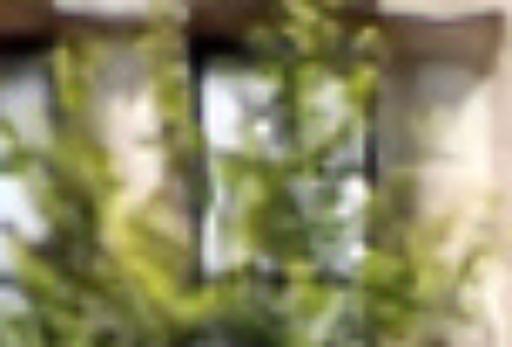}
        \\
        \includegraphics[width=1.0\textwidth]{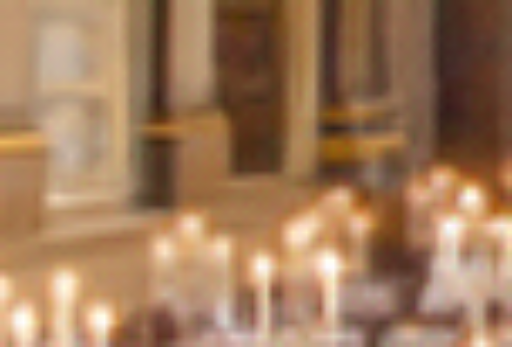}
        \\
        \includegraphics[width=1.0\textwidth]{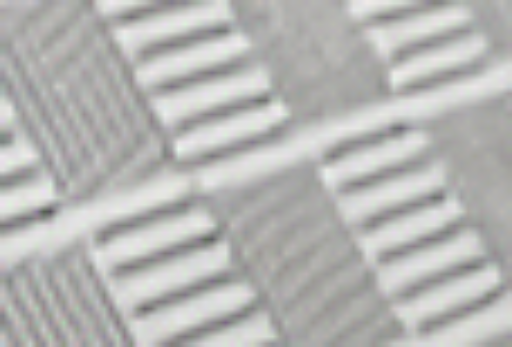}
    \end{minipage}
    }
    \hfill
	\caption{
        Qualitative comparison results using different loss functions on our synthetic DIV2K dataset~\cite{DIV2K} with strong misalignment. 
	}
    \vspace{-10pt}
	\label{fig:DIV2k_comparison}
\end{figure*}

\begin{table}[t]
  \centering
    \footnotesize
    \renewcommand\arraystretch{1.0}
    \setlength{\tabcolsep}{2.0pt}
    \begin{tabular}{c|c|ccccc}
    \hline
    \hline
    Model & Loss & PSNR$\uparrow$  & LPIPS$\downarrow$ & DISTS$\downarrow$ & SSIM$\uparrow$  & FID$\downarrow$ \\
    \hline
    \multirow{4}[2]{*}{NAFNet} & CTX & \textcolor[rgb]{ 0,  .439,  .753}{24.615} & \textcolor[rgb]{ 1,  0,  0}{\textbf{0.245}} & \textcolor[rgb]{ 0,  .439,  .753}{0.105} & \textcolor[rgb]{ 0,  .439,  .753}{0.833} & \textcolor[rgb]{ 0,  .439,  .753}{15.977}  \\
          & LPIPS(Alex) & 16.968 & 0.441 & 0.274 & 0.461 & 35.440 \\
          & PDL   & 17.737 & 0.267 & 0.134 & 0.595 & 16.038 \\
          & \textbf{FDL (Ours)} & \textcolor[rgb]{ 1,  0,  0}{\textbf{24.865}} &\textcolor[rgb]{ 0,  .439,  .753}{ \textbf{0.265}} & \textcolor[rgb]{ 1,  0,  0}{\textbf{0.100}} & \textcolor[rgb]{ 1,  0,  0}{\textbf{0.834}} & \textcolor[rgb]{ 1,  0,  0}{\textbf{15.233}} \\
    \hline
    \multirow{4}[2]{*}{SwinIR} & CTX & \textcolor[rgb]{ 0,  .439,  .753}{35.249} & {0.093} & {0.101} & \textcolor[rgb]{ 0,  .439,  .753}{0.964} & {66.114} \\
          & LPIPS(Alex) & 35.198     & 0.114     & 0.114     & 0.958     & 49.362 \\
          & PDL   & 34.733     & \textcolor[rgb]{ 0,  .439,  .753}{0.086}     & \textcolor[rgb]{ 0,  .439,  .753}{0.094}     & 0.953     & \textcolor[rgb]{ 0,  .439,  .753}{35.275} \\
          & \textbf{FDL (Ours)} & \textcolor[rgb]{ 1,  0,  0}{\textbf{35.771}} & \textcolor[rgb]{ 1,  0,  0}{\textbf{0.085}} & \textcolor[rgb]{ 1,  0,  0}{\textbf{0.088}} & \textcolor[rgb]{ 1,  0,  0}{\textbf{0.965}} & \textcolor[rgb]{ 1,  0,  0}{\textbf{23.299}} \\
    \hline
    \hline
    \end{tabular}%
    \caption{
    Quantitative comparison of SISR on the merged real-world dataset~\cite{cai2019toward, chen2019camera}.}
  \label{tab:RealSR}%
\end{table}%

\subsection{Super Resolution}
We compare our proposed FDL against state-of-the-art loss functions in real-world SISR.
Table~\ref{tab:RealSR} presents the quantitative results of two representative models (\textit{i.e.}, NAFNet and SwinIR), we can observe that our method outperforms all competing methods almost on all evaluation metrics.
On the one hand, our method significantly outperforms the PDL that computes distribution distance in the spatial domain, demonstrating the reasonability of the use of frequency components in our FDL.
On the other hand, our method eliminates the effects of misalignment by utilizing the global structural information in the frequency domain, which still outperforms the CTX.
Furthermore, Table~\ref{tab:DIV2K} reports the comparison results of various loss functions on our synthetic shifted DIV2K dataset with strongly misaligned data.
We can clearly observe that our proposed FDL outperforms all competing loss functions by large margins over all four testing set.
Compared with CTX, our FDL achieves a substantial improvement in PSNR, increasing from 25.14dB to 26.70dB on the Urban100 test set, while also excelling in perceptual metrics such as LPIPS, DISTS, and FID.
This shows that our method achieves a good trade-off between fidelity and quality by measuring distribution distances in the frequency domain.
Figure~\ref{fig:DIV2k_comparison} shows the qualitative results of our FDL and
the other state-of-the-art methods on the synthetic DIV2K dataset.
It is clearly found that the quality of the prediction results of our proposed method is significantly better than that of the comparison method because the results of our method contain less noise and disordered structures.

\begin{table}[t]
    \centering
    \footnotesize
    \renewcommand\arraystretch{1.0}
    \setlength{\tabcolsep}{2.0pt}
    \begin{tabular}{c|c|ccccc}
    \hline
    \hline
    Test Set & Loss & PSNR$\uparrow$ & LPIPS$\downarrow$ & DISTS$\downarrow$ & SSIM$\uparrow$  & FID$\downarrow$  \\
    \hline
    \multirow{4}[2]{*}{Set5} & CTX & \textcolor[rgb]{ 0,  .439,  .753}{30.023} & \textcolor[rgb]{ 0,  .439,  .753}{0.095} & \textcolor[rgb]{ 1,  0,  0}{0.092} & \textcolor[rgb]{ 0,  .439,  .753}{0.933} & \textcolor[rgb]{ 0,  .439,  .753}{5.550} \\
          & LPIPS(Alex) & 21.754 & 0.400   & 0.312 & 0.438 & 66.619 \\
          & PDL   & 29.598  & 0.114  & 0.095  & 0.767  & 7.682  \\
          & \textbf{FDL (Ours)} & \textcolor[rgb]{ 1,  0,  0}{\textbf{32.478}} & \textcolor[rgb]{ 1,  0,  0}{\textbf{0.092}} & \textcolor[rgb]{ 0,  .439,  .753}{\textbf{0.093}} & \textcolor[rgb]{ 1,  0,  0}{\textbf{0.950}} & \textcolor[rgb]{ 1,  0,  0}{\textbf{3.853}} \\
    \hline
    \multirow{4}[2]{*}{Set14} & CTX & \textcolor[rgb]{ 0,  .439,  .753}{27.836} & \textcolor[rgb]{ 0,  .439,  .753}{0.156} & \textcolor[rgb]{ 0,  .439,  .753}{0.105} & \textcolor[rgb]{ 0,  .439,  .753}{0.938} & \textcolor[rgb]{ 0,  .439,  .753}{7.220} \\
          & LPIPS(Alex) & 21.019 & 0.401 & 0.320  & 0.423 & 50.127 \\
          & PDL   & 26.832  & 0.165  & 0.118  & 0.702  & 9.727  \\
          & \textbf{FDL (Ours)} & \textcolor[rgb]{ 1,  0,  0}{\textbf{29.526}} & \textcolor[rgb]{ 1,  0,  0}{\textbf{0.152}} & \textcolor[rgb]{ 1,  0,  0}{\textbf{0.103}} & \textcolor[rgb]{ 1,  0,  0}{\textbf{0.957}} & \textcolor[rgb]{ 1,  0,  0}{\textbf{5.853}} \\
    \hline
    \multirow{4}[2]{*}{B100} & CTX & \textcolor[rgb]{ 0,  .439,  .753}{27.829} & \textcolor[rgb]{ 0,  .439,  .753}{0.152} & \textcolor[rgb]{ 0,  .439,  .753}{0.114} & \textcolor[rgb]{ 0,  .439,  .753}{0.881} & \textcolor[rgb]{ 1,  0,  0}{16.681} \\
          & LPIPS(Alex) & 22.041 & 0.367 & 0.299 & 0.390  & 91.150 \\
          & PDL   & 27.231  & 0.159  & 0.123  & 0.645  & \textcolor[rgb]{ 0,  .439,  .753}{17.308 } \\
          & \textbf{FDL (Ours)} & \textcolor[rgb]{ 1,  0,  0}{\textbf{28.968}} & \textcolor[rgb]{ 1,  0,  0}{\textbf{0.150}} & \textcolor[rgb]{ 1,  0,  0}{\textbf{0.110}} & \textcolor[rgb]{ 1,  0,  0}{\textbf{0.902}} & \textbf{20.451} \\
    \hline
    \multirow{4}[2]{*}{Urban100} & CTX & \textcolor[rgb]{ 0,  .439,  .753}{25.138} & \textcolor[rgb]{ 0,  .439,  .753}{0.143} & \textcolor[rgb]{ 0,  .439,  .753}{0.104} & \textcolor[rgb]{ 0,  .439,  .753}{0.850} & \textcolor[rgb]{ 0,  .439,  .753}{8.582} \\
          & LPIPS(Alex) & 19.987 & 0.382 & 0.297 & 0.369 & 79.639 \\
          & PDL   & 23.847  & 0.194  & 0.136  & 0.572  & 22.389  \\
          & \textbf{FDL (Ours)} & \textcolor[rgb]{ 1,  0,  0}{\textbf{26.702}} & \textcolor[rgb]{ 1,  0,  0}{\textbf{0.137}} & \textcolor[rgb]{ 1,  0,  0}{\textbf{0.098}} & \textcolor[rgb]{ 1,  0,  0}{\textbf{0.887}} & \textcolor[rgb]{ 1,  0,  0}{\textbf{7.903}} \\
    \hline
    \hline
    \end{tabular}%
    \caption{
    Quantitative comparison of NLSN~\cite{mei2021image} for SISR on our synthetic shifted DIV2K dataset~\cite{DIV2K}.
    }
  \label{tab:DIV2K}%
\end{table}%

\subsection{Style Transfer}
Style transfer aims to synthesize a new image that combines the content of one image with the artistic or stylistic features of another.
The primary challenge of style transfer is finding the right balance between preserving the content of the input image and incorporating the stylistic features from the reference style image. 
Following the pipeline of Gatys \etal~\cite{style_transfer}, we optimize the generated image with content loss and style loss. 
Our proposed FDL is also capable of handling this challenging task, since the use of distribution distance measurement in the frequency domain.
Specifically, we define content loss and style loss as follows:
\begin{equation}
    \mathcal{L}_{\text{style}}(R,S) = \mathcal{L}_{\text{FDL}}\left(R,S \right),
    \label{style_loss}
\end{equation}
\begin{equation}
    \mathcal{L}_{\text{content}}(R,T) = \text{SW}\left(\mathcal{P}_{\Phi(R)},\mathcal{P}_{\Phi(T)} \right),
    \label{content_loss}
\end{equation}
where $R$ is the generated image, $S$ and $T$ refer to style and content image, respectively.
We compare FDL with CTX and perceptual losses in Gatys \etal, and use their respective official settings for fair comparison.

\begin{figure}[t]
    \centering 
    \subfloat[Content]{
    \begin{minipage}[b]{0.09\textwidth}
        \includegraphics[width=1.0\textwidth]{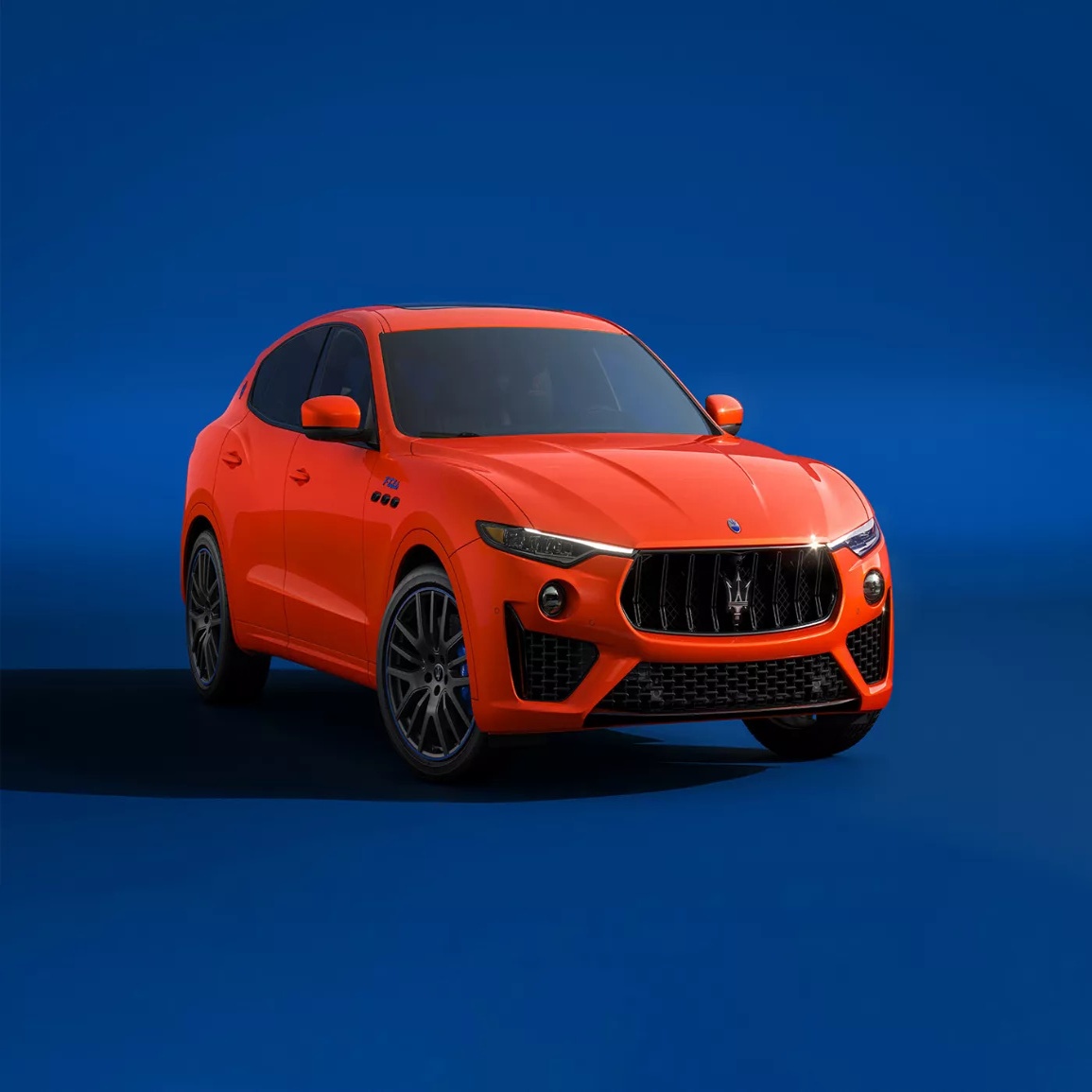}
        \\
        \includegraphics[width=1.0\textwidth]{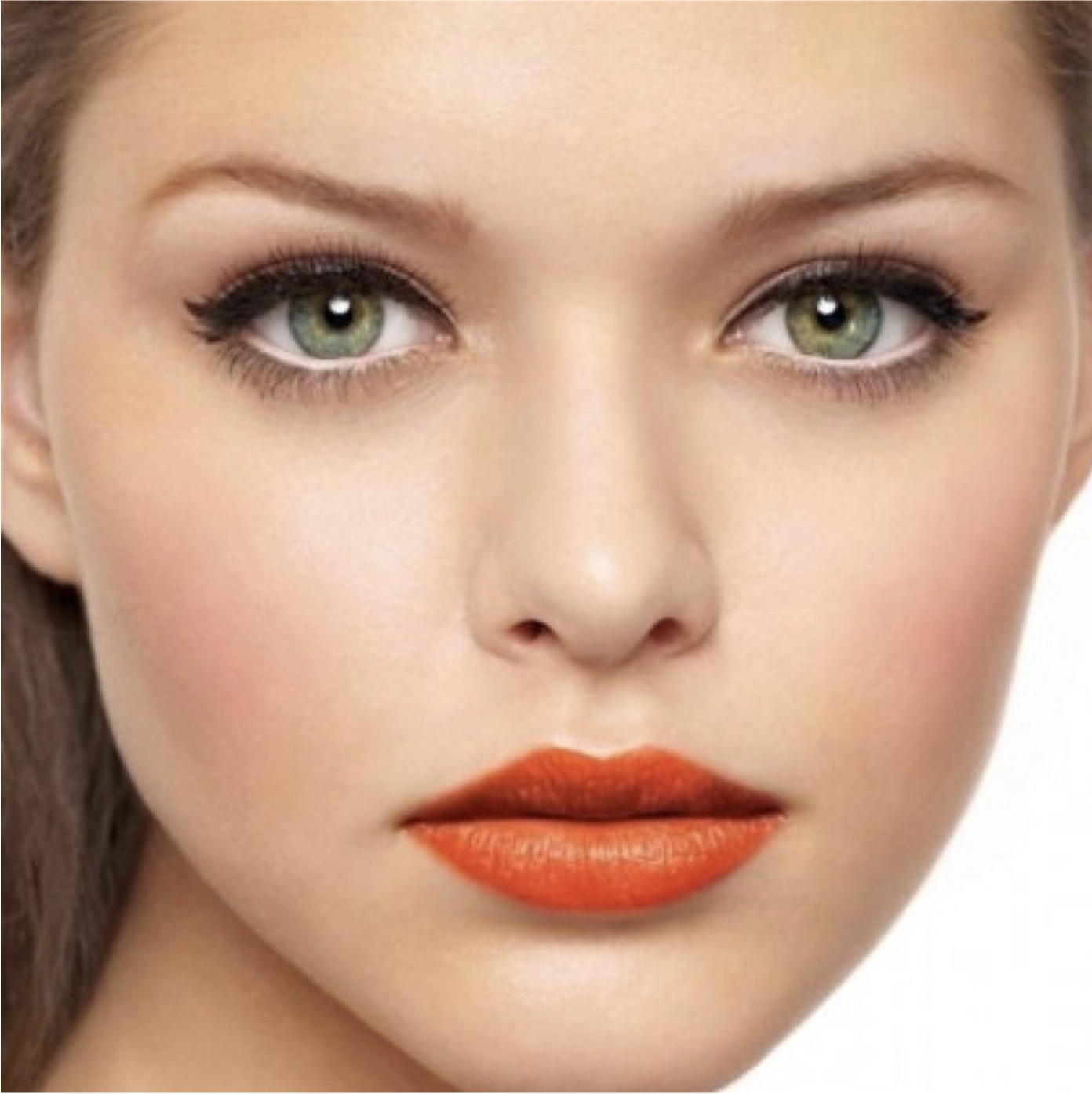}
        \\
        \includegraphics[width=1.0\textwidth]{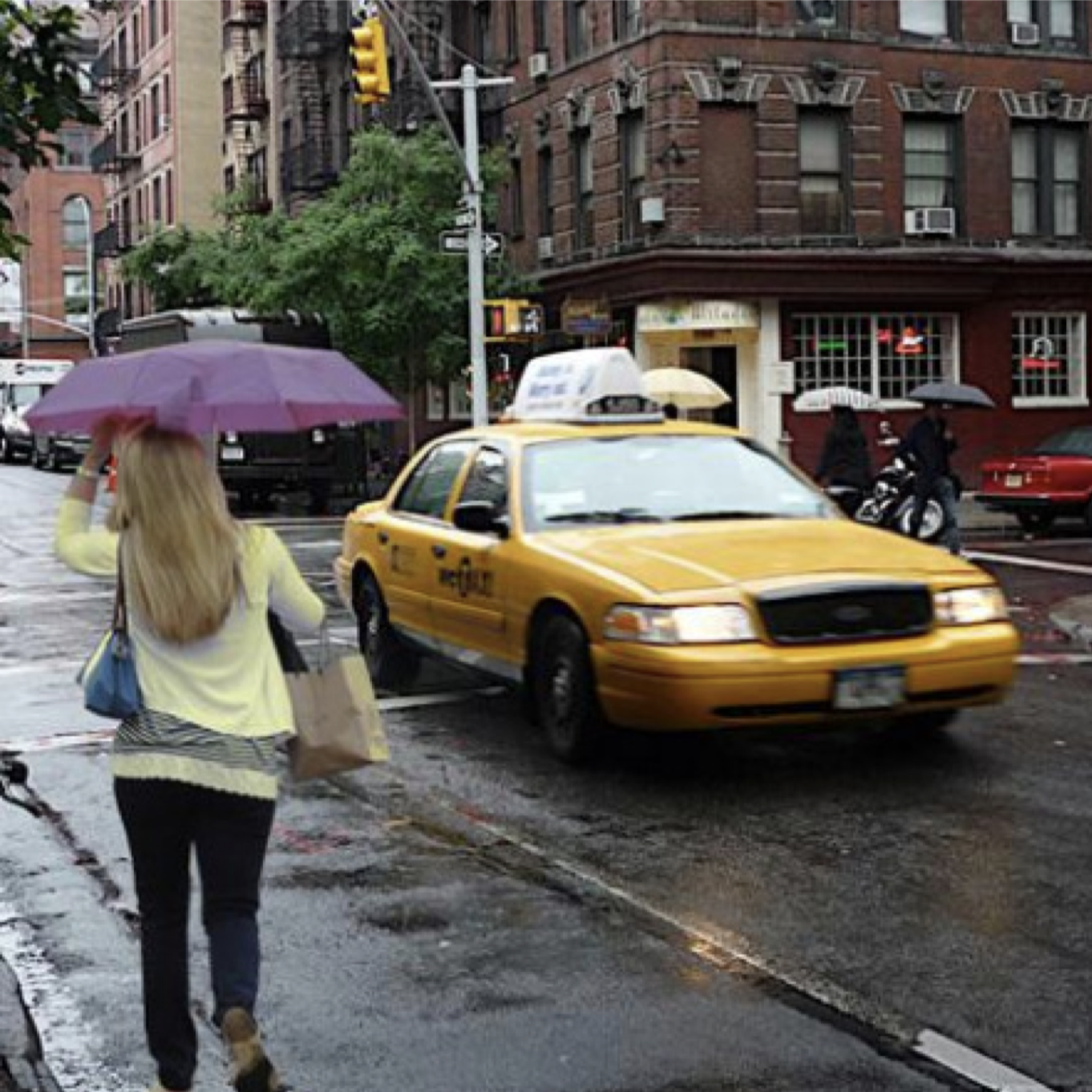}
        \\
        \includegraphics[width=1.0\textwidth]{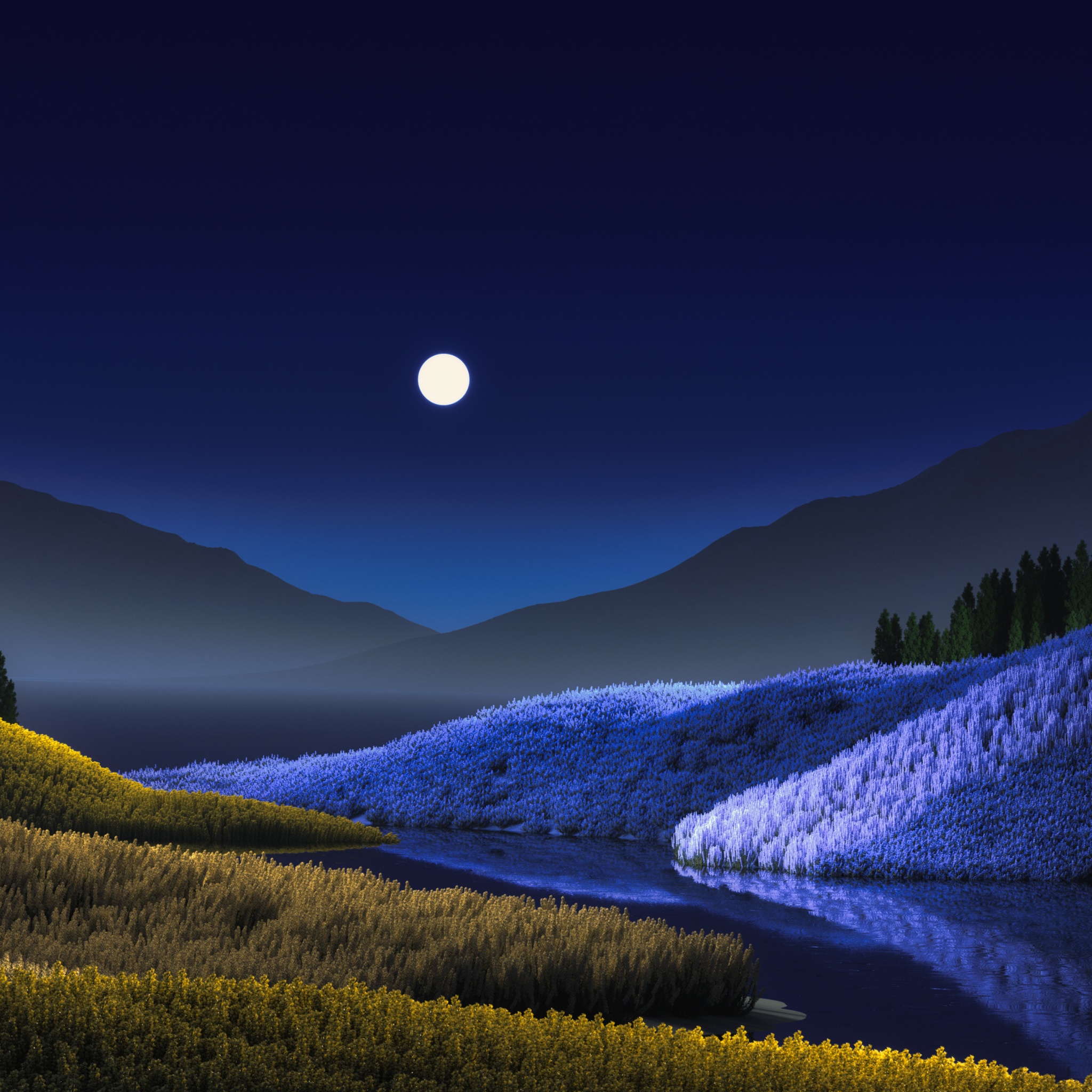}
    \end{minipage} 
    } 
    \subfloat[Style]{
    \begin{minipage}[b]{0.09\textwidth}
        \includegraphics[width=1.0\textwidth]{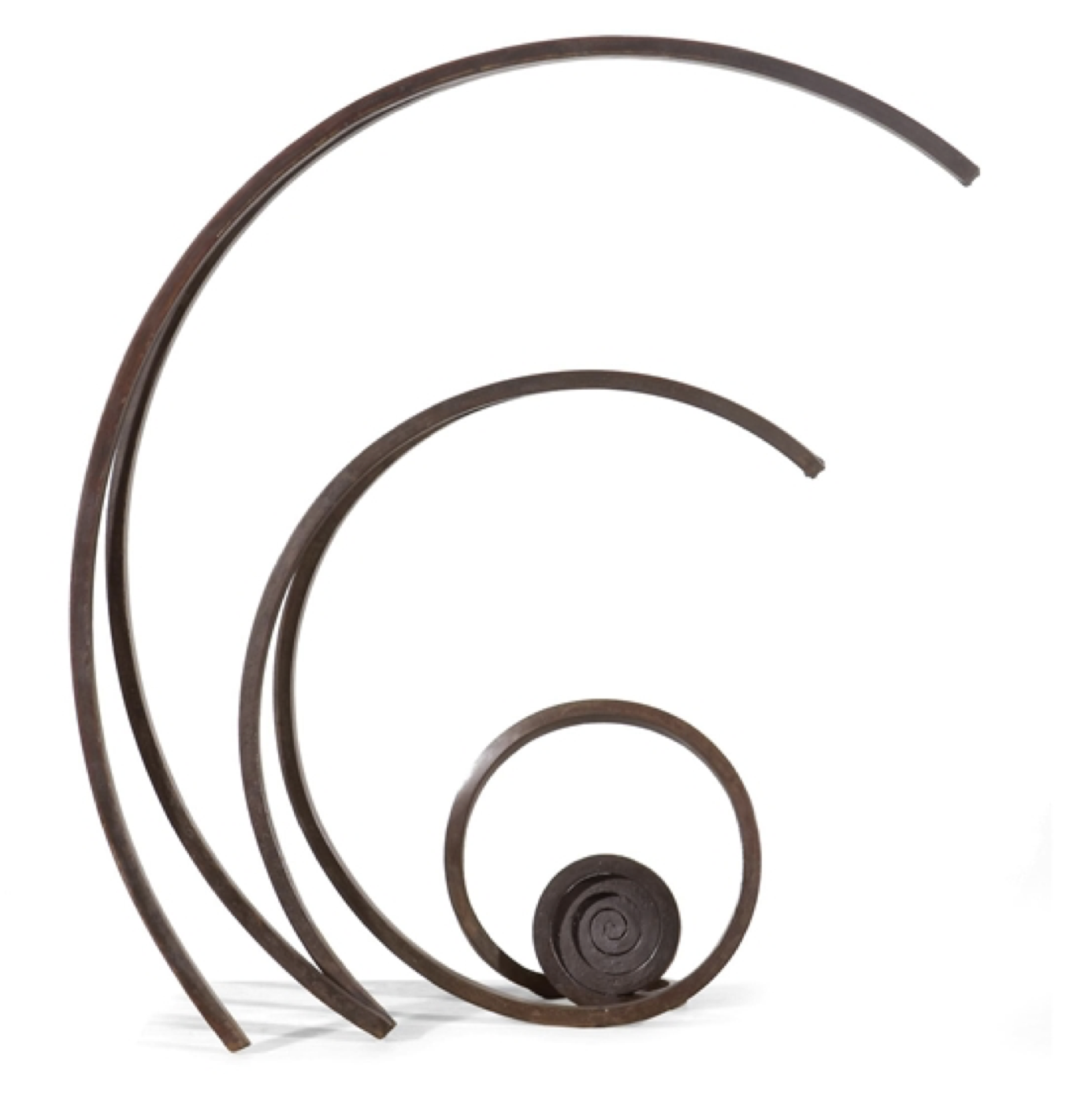}
        \\
        \includegraphics[width=1.0\textwidth]{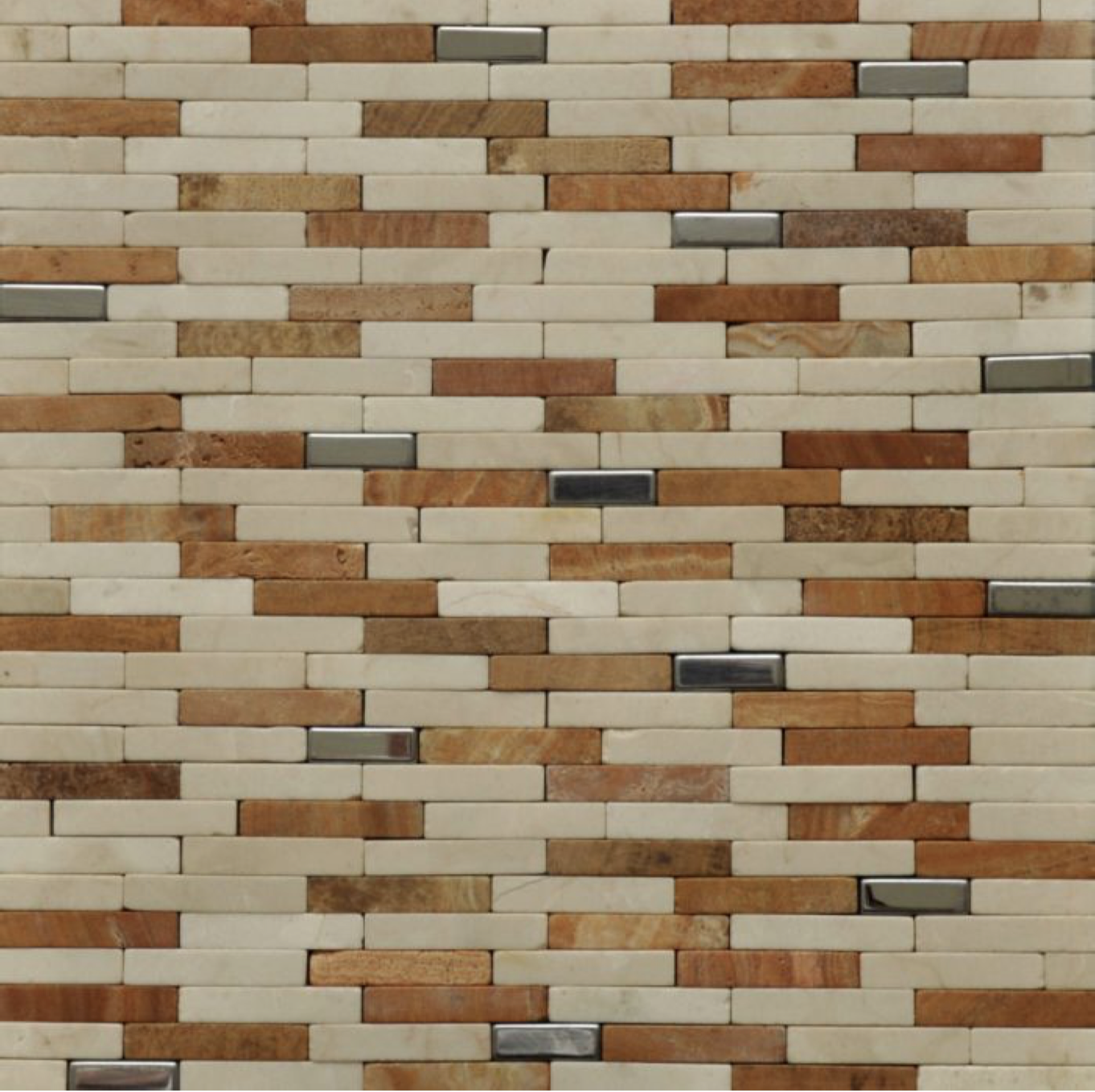}
        \\
        \includegraphics[width=1.0\textwidth]{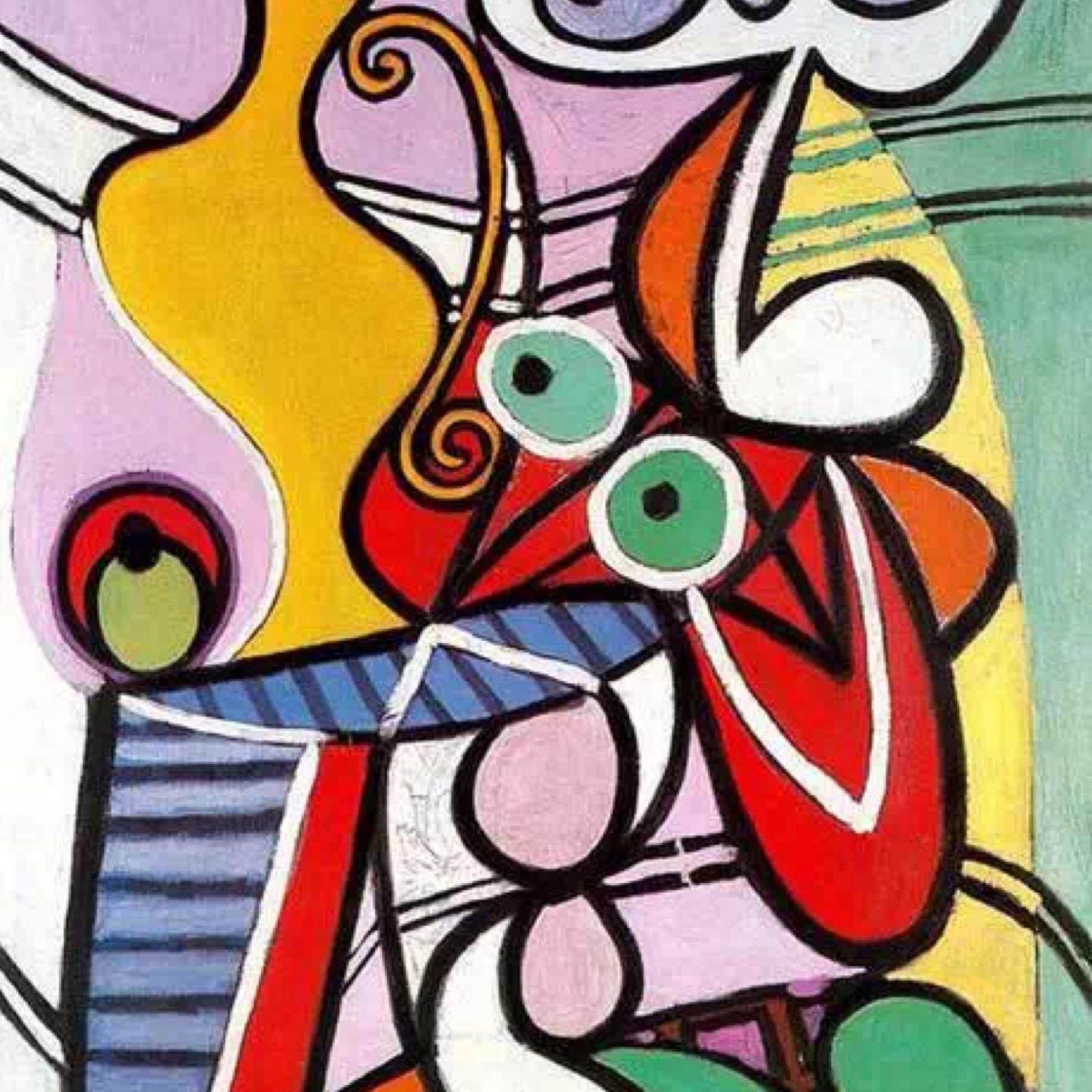}
        \\
        \includegraphics[width=1.0\textwidth]{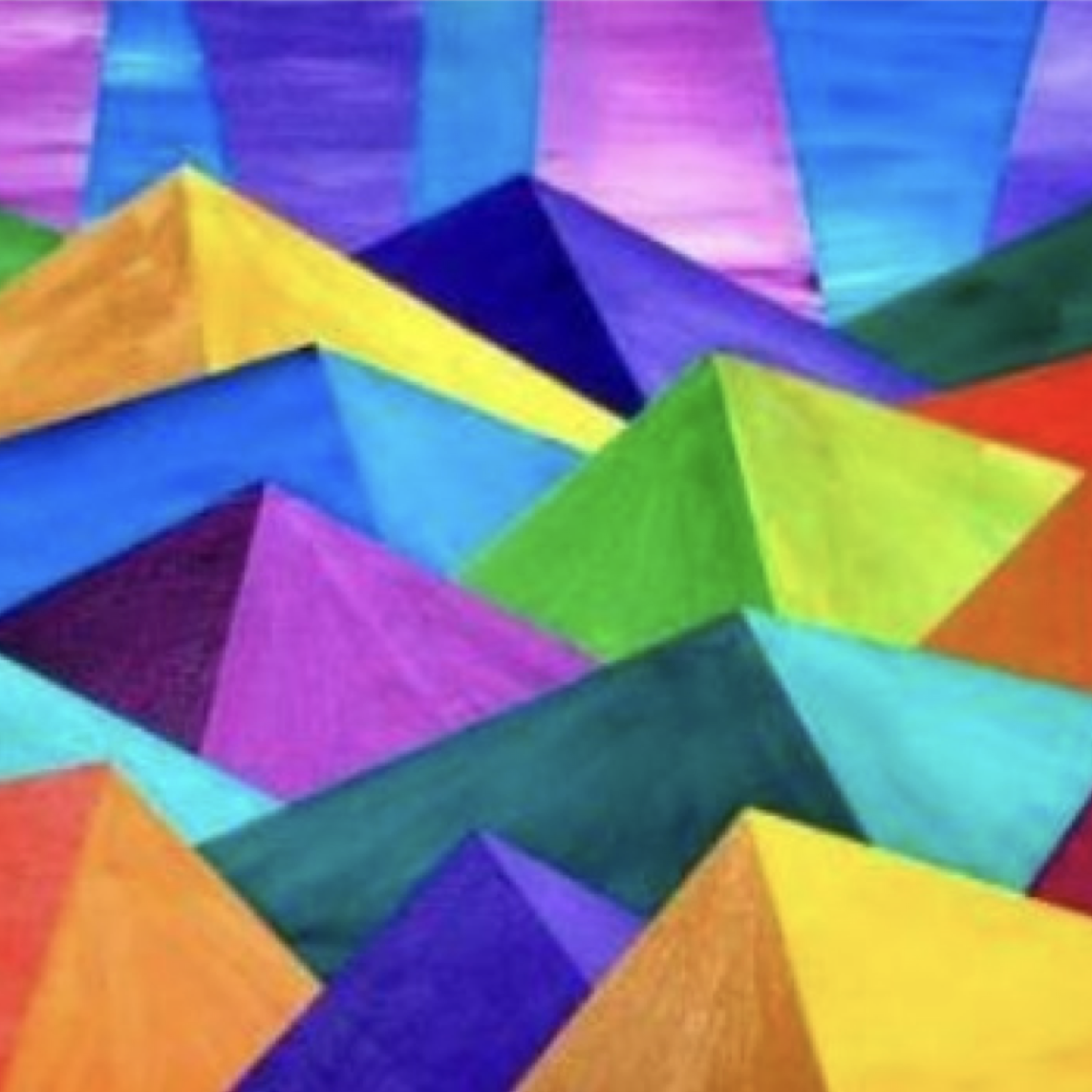}
    \end{minipage}
    }
    \subfloat[Gatys \etal]{
    \begin{minipage}[b]{0.09\textwidth}
        \includegraphics[width=1.0\textwidth]{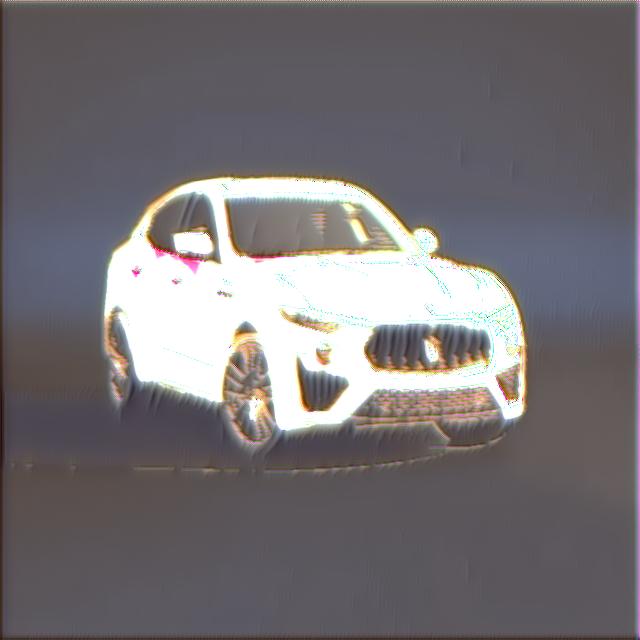}
        \\
        \includegraphics[width=1.0\textwidth]{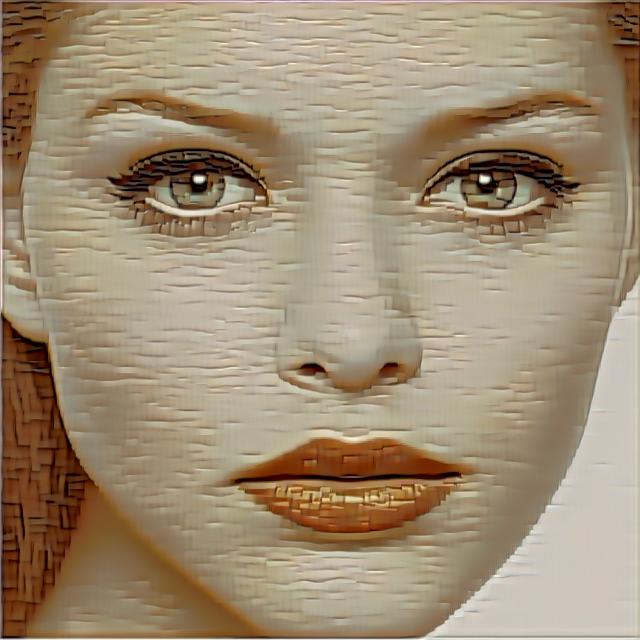}
        \\
        \includegraphics[width=1.0\textwidth]{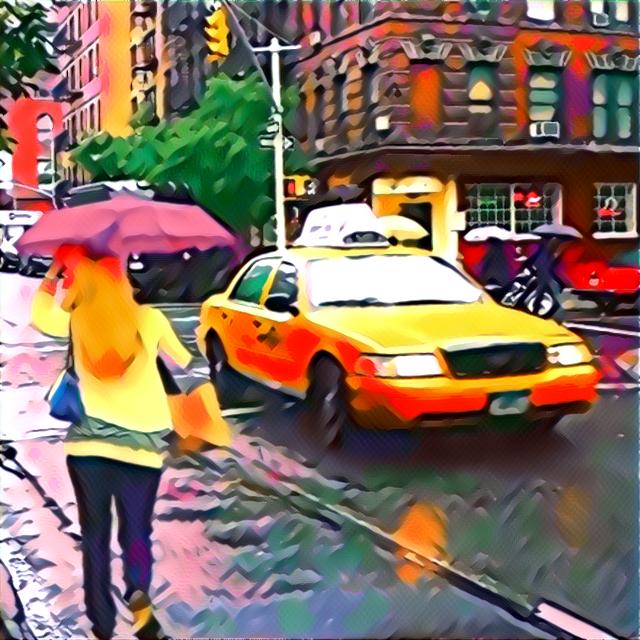}
        \\
        \includegraphics[width=1.0\textwidth]{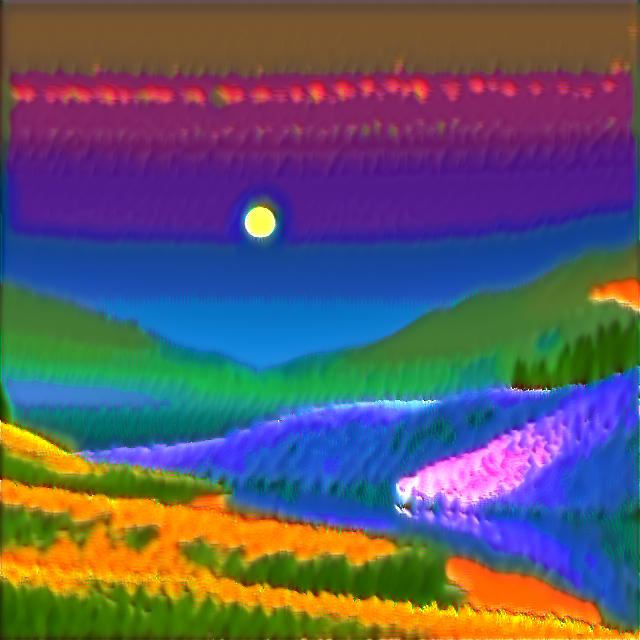}
    \end{minipage}
    }
    \subfloat[CTX]{
    \begin{minipage}[b]{0.09\textwidth}
        \includegraphics[width=1.0\textwidth]{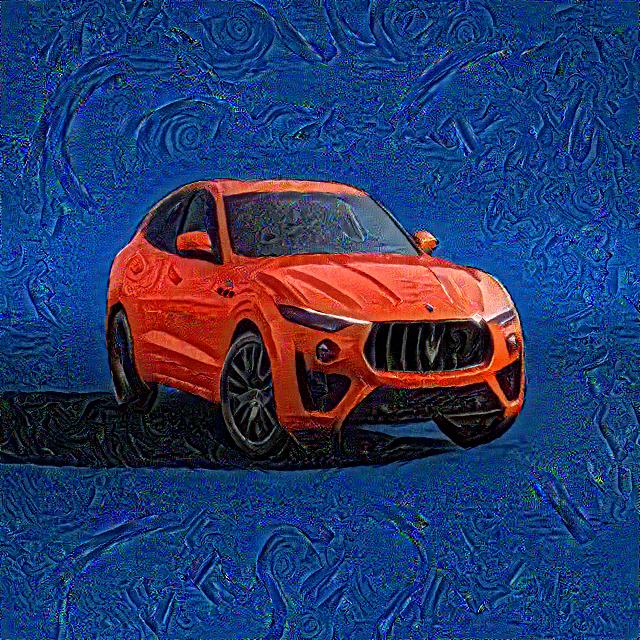}
        \\
        \includegraphics[width=1.0\textwidth]{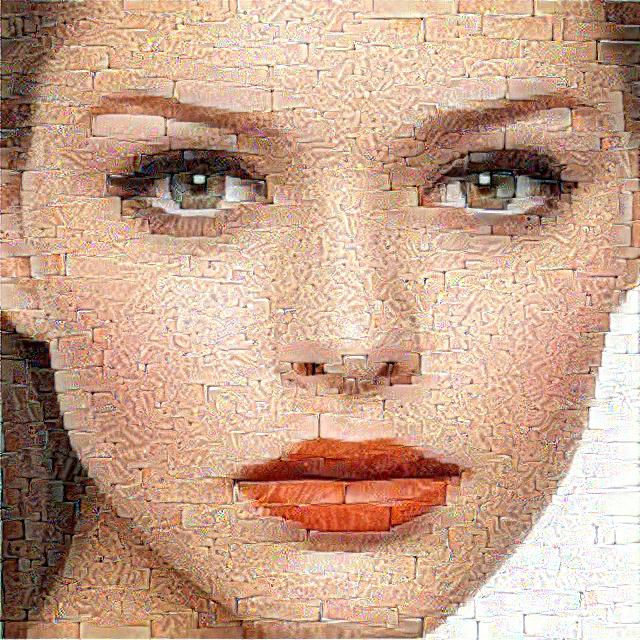}
        \\
        \includegraphics[width=1.0\textwidth]{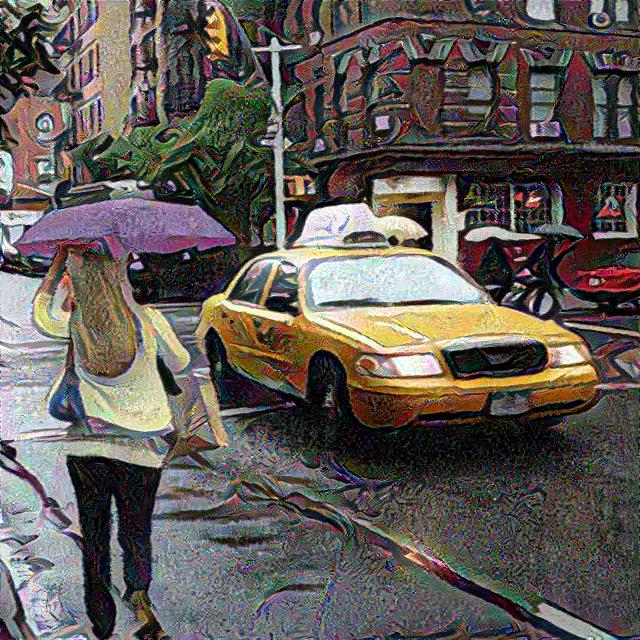}
        \\
        \includegraphics[width=1.0\textwidth]{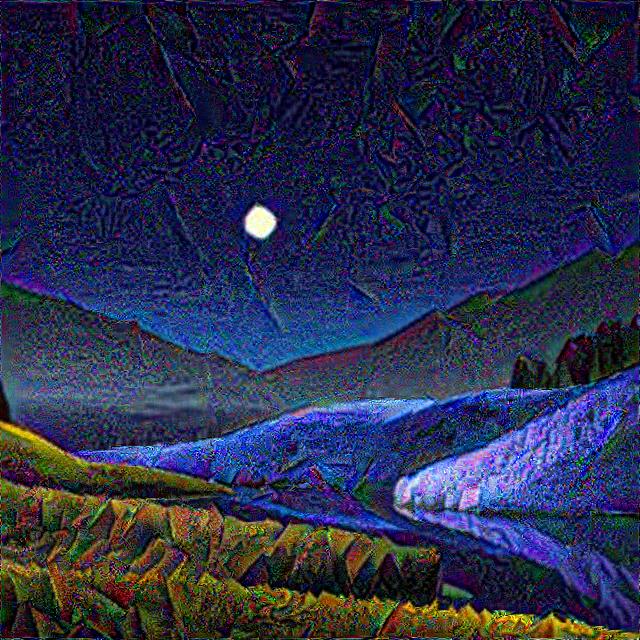}
    \end{minipage}
    }
    \subfloat[FDL (Ours)]{
    \begin{minipage}[b]{0.09\textwidth}
        \includegraphics[width=1.0\textwidth]{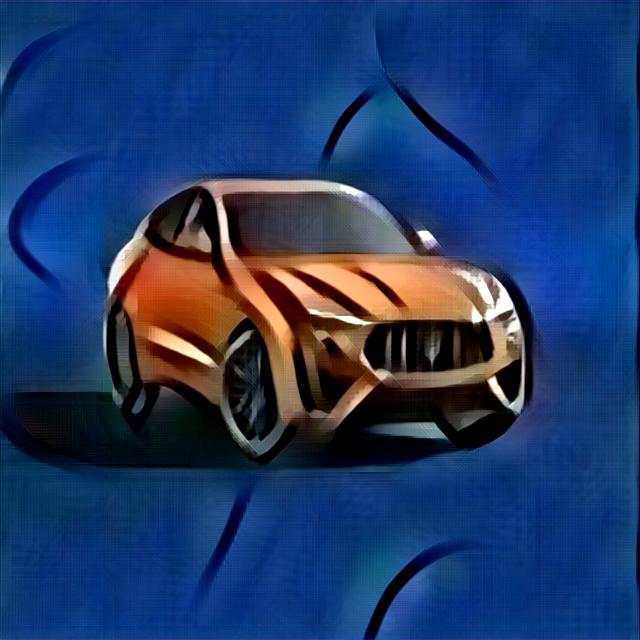}
        \\
        \includegraphics[width=1.0\textwidth]{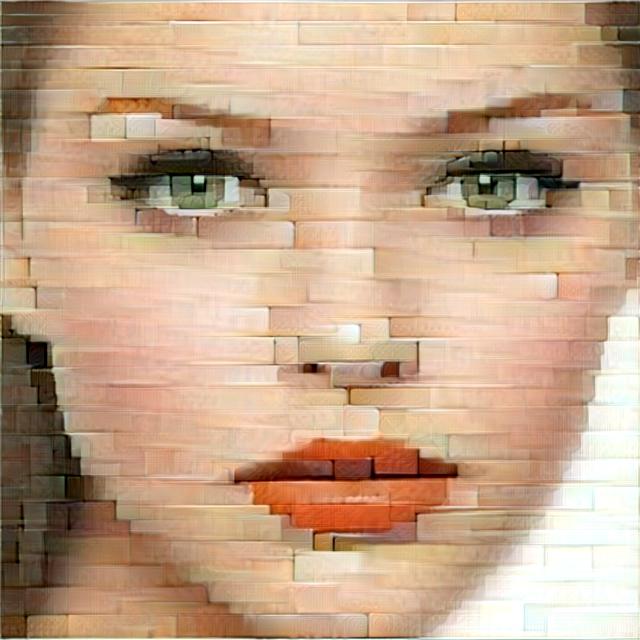}
        \\
        \includegraphics[width=1.0\textwidth]{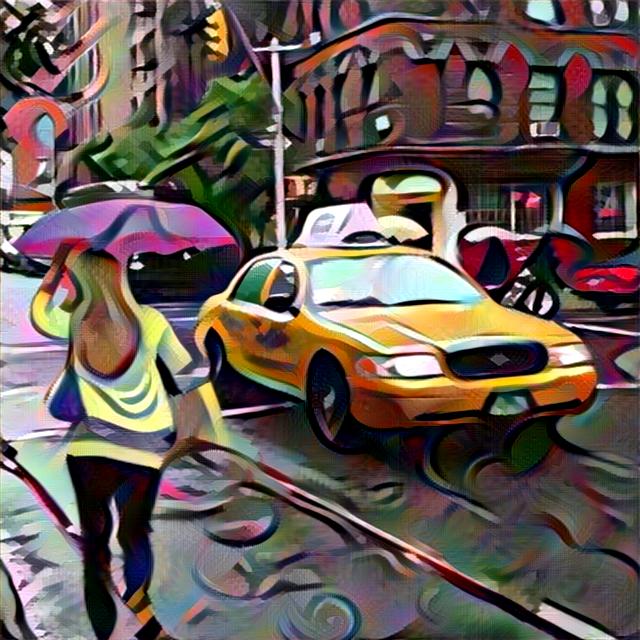}
        \\
        \includegraphics[width=1.0\textwidth]{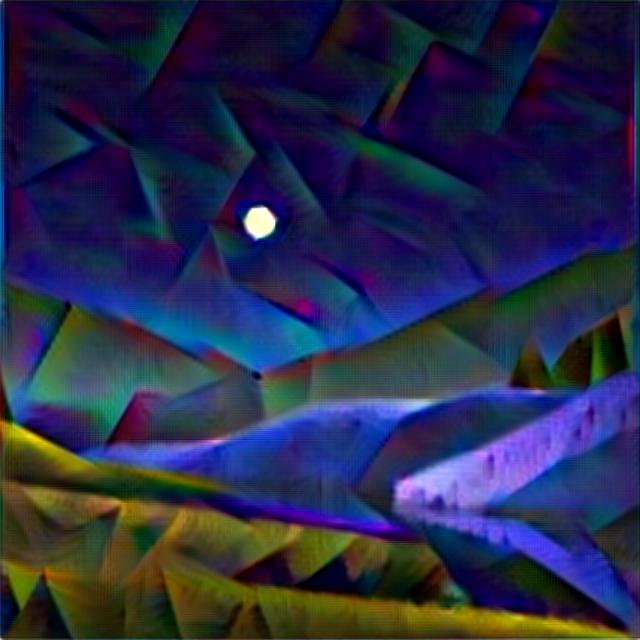}
    \end{minipage}
    }
    \hfill
	\caption{
        Qualitative comparison results compared with Gatys \etal and CTX. Our FDL loss function can better retain the structural information in style images. 
	}
	\label{fig:style_comparison}
\end{figure}

Visual comparison results are presented in Figure~\ref{fig:style_comparison}. It can be observed that losses in Gatys \etal~\cite{style_transfer} only capture the color information from the style image, resulting in poor performance in transferring structured styles. 
On the other hand, CTX focuses on local texture patterns in the style image but fails to achieve style transfer at a more global and structural level. 
In contrast, our method can effectively capture the structural information present in the style image.
This demonstrates the effectiveness of utilizing frequency domain global information.

\begin{table}[t]
    \centering
    \footnotesize
    \renewcommand\arraystretch{1.0}
    \setlength{\tabcolsep}{6.0pt}
    \begin{tabular}{c|cccccc}
    \hline
    \hline
    Loss  & PSNR$\uparrow$ & LPIPS$\downarrow$ & DISTS$\downarrow$ & SSIM$\uparrow$  & FID$\downarrow$ \\
    \hline
    $\mathcal{L}_{\text{spatial}}$ & 22.916 & 0.118 & 0.125 & 0.798 & 61.087 \\
    \hline
    $\mathcal{L}_{\text{FDL}}$  & \textcolor[rgb]{ 1,  0,  0}{23.048} & \textcolor[rgb]{ 1,  0,  0}{0.114} & \textcolor[rgb]{ 1,  0,  0}{0.121} & \textcolor[rgb]{ 1,  0,  0}{0.811} & \textcolor[rgb]{ 1,  0,  0}{37.501} \\
    \hline
    \hline
    \end{tabular}%
    \caption{Ablation of calculating in frequency domain.}
    \vspace{-10pt}
  \label{tab:ablation_freq}%
\end{table}%

\subsection{Ablation Study}
We conduct a series of ablation experiments on the NAFNet in the DPED dataset.
Firstly, to validate the impact of computing distribution distance in the frequency domain, we calculate the SWD~\cite{patchSWD} as the loss function in the image feature spatial domain, as follows:
\begin{equation}
\mathcal{L}_{\text{spatial}}(U,V)=\text{SW}\left(\Phi(U),\Phi(V) \right).
\label{spatial_SWD}
\end{equation}
From Table~\ref{tab:ablation_freq}, compared with $\mathcal{L}_{\text{spatial}}$, we can observe that the proposed FDL shows improvements across all metrics. This observation suggests that computing the distribution distance between global information in the frequency domain as a loss function can better ensure the overall quality of the predicted results.

\begin{table}[t]
    \centering
    \footnotesize
    \renewcommand\arraystretch{1.0}
    \setlength{\tabcolsep}{1.6pt}
    \begin{tabular}{c|c|ccccc}
    \hline
    \hline
    Loss  & Backbone & PSNR$\uparrow$ & LPIPS$\downarrow$ & DISTS$\downarrow$ & SSIM$\uparrow$  & FID$\downarrow$ \\
    \hline
    $\mathcal{L}_{\text{FDL}}$($\lambda=0.01$)   & ResNet & 22.415 & 0.139 & 0.146 & 0.763 & 79.812 \\
    $\mathcal{L}_{\text{FDL}}$($\lambda=0.01$)   & EffNet & 20.389 & 0.196 & 0.178 & 0.578 & 173.891 \\
    $\mathcal{L}_{\text{FDL}}$($\lambda=0.01$)   & None &  22.581 & 0.149 & 0.156 & 0.789 & 60.110 \\
    \hline
    $\mathcal{L}_{\text{FDL}}$($\lambda=100$)   & VGG   & 22.134 & 0.131 & 0.141 & 0.787 & 67.766 \\
    $\mathcal{L}_{\text{FDL}}$($\lambda=10$)   & VGG   & 22.815 & 0.121 & 0.128 & 0.803 & 52.503 \\
    $\mathcal{L}_{\text{FDL}}$($\lambda=1$)   & VGG   & 22.810 & 0.117 & 0.124 & 0.806 & 55.629 \\
    $\mathcal{L}_{\text{FDL}}$($\lambda=0.1$)   & VGG   & 22.822 & 0.118 & 0.126 & 0.804 & 65.711 \\
\hline 
    $\mathcal{L}_{\text{FDL}}$($\lambda=0.01$)   & VGG   & \textcolor[rgb]{ 1,  0,  0}{23.048} & \textcolor[rgb]{ 1,  0,  0}{0.114} & \textcolor[rgb]{ 1,  0,  0}{0.121} & \textcolor[rgb]{ 1,  0,  0}{0.811} & \textcolor[rgb]{ 1,  0,  0}{37.501} \\
    \hline
    \hline
    \end{tabular}%
    \caption{Ablation of backbone feature extractor and the weight of SWD between phase component of features.}
  \label{tab:ablation}%
\end{table}%

Next, we aim to further investigate the effect of different feature extractors on the final results. 
To achieve this, we conducted experiments by using ResNet~\cite{he2016deep} and EffNet~\cite{freeman2018effnet} as feature extractors in our proposed loss. 
In particular, we also remove the feature extractor in the proposed loss function, thereby directly performing FDL loss calculation on pixels (i.e., the "None" in Table~\ref{tab:ablation}).
The quantitative results of different feature extractors are shown in Table~\ref{tab:ablation}, and we can observe that VGG19~\cite{Simonyan2014VeryDC} yields the best performance among all the results.

Finally, we aim to explore the impact of the weight assigned to the distribution distance between amplitude and phase, by adjusting $\lambda$ in Equation~\ref{eq:FDL}.
Table~\ref{tab:ablation} reports the comparison results of different settings of $\lambda$, and we can observe that $\lambda=0.01$ performs best among all settings. 
This can be attributed to the fact that in the DPED~\cite{ignatovDSLRQualityPhotosMobile2017a} dataset, the main difference between the input and target images lies in the texture, which is highly correlated with the amplitude of the image features.
Therefore, assigning a higher weight to the amplitude component in FDL helps the model achieve better performance on the DPED dataset.
This observation suggests that in image transformation tasks with different emphasis on different image characteristics, adjusting the value of $\lambda$ allows the model to allocate different priorities to the characteristics of the predicted results.

\section{Conclusion}
\label{sec:conclusion}

%
%

This paper proposes a robust misalignment loss for image transformation tasks. 
Our proposed FDL calculates the distribution distance in the frequency domain of image features.
Through experiments, we have demonstrated that frequency domain components of image features contain global information closely related to multiple image characteristics.
By utilizing the distance between the distribution of these global information as a loss function, we can mitigate the limitation in spatial distribution distances, and ensure the overall quality of the predicted results.
In future work, we hope to investigate the frequency components of image features further and improve the performance of FDL by assigning different attention weights to distinct frequency domain regions.

\section{Acknowledgments}
\label{sec:acknowledgments}

This work was supported in part by the National Natural Science Foundation of China under Grant 62201387 and 62371343, in part by the Shanghai Pujiang Program under Grant 22PJ1413300, and in part by the Fundamental Research Funds for the Central Universities. 

{
    \small
    \bibliographystyle{ieeenat_fullname}
    \bibliography{main}

\begin{thebibliography}{45}
\providecommand{\natexlab}[1]{#1}
\providecommand{\url}[1]{\texttt{#1}}
\expandafter\ifx\csname urlstyle\endcsname\relax
  \providecommand{\doi}[1]{doi: #1}\else
  \providecommand{\doi}{doi: \begingroup \urlstyle{rm}\Url}\fi

\bibitem[Agustsson and Timofte(2017)]{DIV2K}
Eirikur Agustsson and Radu Timofte.
\newblock {NTIRE 2017 challenge on single image super-resolution: Dataset and
  study}.
\newblock In \emph{Proceedings of the IEEE/CVF Conference on Computer Vision
  and Pattern Recognition Workshops}, 2017.

\bibitem[Blau and Michaeli(2018)]{blauPerceptionDistortionTradeoff2018}
Yochai Blau and Tomer Michaeli.
\newblock The perception-distortion tradeoff.
\newblock In \emph{Proceedings of the IEEE/CVF Conference on Computer Vision
  and Pattern Recognition}, pages 6228--6237, 2018.

\bibitem[Cai et~al.(2019)Cai, Zeng, Yong, Cao, and Zhang]{cai2019toward}
Jianrui Cai, Hui Zeng, Hongwei Yong, Zisheng Cao, and Lei Zhang.
\newblock Toward real-world single image super-resolution: A new benchmark and
  a new model.
\newblock In \emph{Proceedings of the IEEE/CVF International Conference on
  Computer Vision}, pages 3086--3095, 2019.

\bibitem[Cai et~al.(2021)Cai, Zhang, Huang, Geng, Li, and Huang]{FDIT}
Mu Cai, Hong Zhang, Huijuan Huang, Qichuan Geng, Yixuan Li, and Gao Huang.
\newblock Frequency domain image translation: More photo-realistic, better
  identity-preserving.
\newblock In \emph{Proceedings of the IEEE/CVF International Conference on
  Computer Vision}, pages 13930--13940, 2021.

\bibitem[Chen et~al.(2019)Chen, Xiong, Tian, Zha, and Wu]{chen2019camera}
Chang Chen, Zhiwei Xiong, Xinmei Tian, Zheng-Jun Zha, and Feng Wu.
\newblock Camera lens super-resolution.
\newblock In \emph{Proceedings of the IEEE/CVF Conference on Computer Vision
  and Pattern Recognition}, pages 1652--1660, 2019.

\bibitem[Chen et~al.(2022)Chen, Chu, Zhang, and Sun]{chen2022simple}
Liangyu Chen, Xiaojie Chu, Xiangyu Zhang, and Jian Sun.
\newblock Simple baselines for image restoration.
\newblock In \emph{European Conference on Computer Vision}, pages 17--33.
  Springer, 2022.

\bibitem[Creswell et~al.(2018)Creswell, White, Dumoulin, Arulkumaran, Sengupta,
  and Bharath]{creswell2018generative}
Antonia Creswell, Tom White, Vincent Dumoulin, Kai Arulkumaran, Biswa Sengupta,
  and Anil~A Bharath.
\newblock Generative adversarial networks: An overview.
\newblock \emph{IEEE Signal Processing Magazine}, 35\penalty0 (1):\penalty0
  53--65, 2018.

\bibitem[Delbracio et~al.(2021)Delbracio, Talebei, and
  Milanfar]{delbracioProjectedDistributionLoss2021}
Mauricio Delbracio, Hossein Talebei, and Pevman Milanfar.
\newblock Projected distribution loss for image enhancement.
\newblock In \emph{2021 IEEE International Conference on Computational
  Photography}, pages 1--12, 2021.

\bibitem[Ding et~al.(2020)Ding, Ma, Wang, and
  Simoncelli]{dingImageQualityAssessment2020}
Keyan Ding, Kede Ma, Shiqi Wang, and Eero~P Simoncelli.
\newblock Image quality assessment: Unifying structure and texture similarity.
\newblock \emph{IEEE Transactions on Pattern Analysis and Machine
  Intelligence}, 44\penalty0 (5):\penalty0 2567--2581, 2020.

\bibitem[Dong et~al.(2014)Dong, Loy, He, and
  Tang]{dongLearningDeepConvolutional2014}
Chao Dong, Chen~Change Loy, Kaiming He, and Xiaoou Tang.
\newblock Learning a deep convolutional network for image super-resolution.
\newblock In \emph{European Conference on Computer Vision}, pages 184--199.
  Springer, 2014.

\bibitem[Elnekave and Weiss(2022)]{patchSWD}
Ariel Elnekave and Yair Weiss.
\newblock Generating natural images with direct patch distributions matching.
\newblock In \emph{European Conference on Computer Vision}, pages 544--560.
  Springer, 2022.

\bibitem[Freeman et~al.(2018)Freeman, Roese-Koerner, and
  Kummert]{freeman2018effnet}
Ido Freeman, Lutz Roese-Koerner, and Anton Kummert.
\newblock Effnet: An efficient structure for convolutional neural networks.
\newblock In \emph{IEEE International Conference on Image Processing}, pages
  6--10. IEEE, 2018.

\bibitem[Gaspar and Rousselet(2009)]{phase1}
Carl~M Gaspar and Guillaume~A Rousselet.
\newblock How do amplitude spectra influence rapid animal detection?
\newblock \emph{Vision Research}, 49\penalty0 (24):\penalty0 3001--3012, 2009.

\bibitem[Gatys et~al.(2016)Gatys, Ecker, and Bethge]{style_transfer}
Leon~A Gatys, Alexander~S Ecker, and Matthias Bethge.
\newblock Image style transfer using convolutional neural networks.
\newblock In \emph{Proceedings of the IEEE/CVF Conference on Computer Vision
  and Pattern Recognition}, pages 2414--2423, 2016.

\bibitem[Gharbi et~al.(2017)Gharbi, Chen, Barron, Hasinoff, and
  Durand]{gharbi2017deep}
Micha{\"e}l Gharbi, Jiawen Chen, Jonathan~T Barron, Samuel~W Hasinoff, and
  Fr{\'e}do Durand.
\newblock Deep bilateral learning for real-time image enhancement.
\newblock \emph{In ACM Transactions on Graphics}, 36\penalty0 (4):\penalty0
  1--12, 2017.

\bibitem[Ghildyal and Liu(2022)]{ghildyalShiftTolerantPerceptualSimilarity2022}
Abhijay Ghildyal and Feng Liu.
\newblock Shift-tolerant perceptual similarity metric.
\newblock In \emph{European Conference on Computer Vision}, pages 91--107.
  Springer, 2022.

\bibitem[He et~al.(2016)He, Zhang, Ren, and Sun]{he2016deep}
Kaiming He, Xiangyu Zhang, Shaoqing Ren, and Jian Sun.
\newblock Deep residual learning for image recognition.
\newblock In \emph{Proceedings of the IEEE/CVF Conference on Computer Vision
  and Pattern Recognition}, pages 770--778, 2016.

\bibitem[Heitz et~al.(2021)Heitz, Vanhoey, Chambon, and
  Belcour]{heitz2021sliced}
Eric Heitz, Kenneth Vanhoey, Thomas Chambon, and Laurent Belcour.
\newblock A sliced {Wasserstein} loss for neural texture synthesis.
\newblock In \emph{Proceedings of the IEEE/CVF Conference on Computer Vision
  and Pattern Recognition}, pages 9412--9420, 2021.

\bibitem[Heusel et~al.(2017)Heusel, Ramsauer, Unterthiner, Nessler, and
  Hochreiter]{heusel2017gans}
Martin Heusel, Hubert Ramsauer, Thomas Unterthiner, Bernhard Nessler, and Sepp
  Hochreiter.
\newblock {GANs} trained by a two time-scale update rule converge to a local
  nash equilibrium.
\newblock \emph{Advances in Neural Information Processing Systems}, 30, 2017.

\bibitem[Huang et~al.(2022)Huang, Liu, Zhao, Yan, Zhang, Huang, Zhou, and
  Xiong]{huangDeepFourierBasedExposure2022}
Jie Huang, Yajing Liu, Feng Zhao, Keyu Yan, Jinghao Zhang, Yukun Huang, Man
  Zhou, and Zhiwei Xiong.
\newblock Deep fourier-based exposure correction network with spatial-frequency
  interaction.
\newblock In \emph{European Conference on Computer Vision}, pages 163--180.
  Springer, 2022.

\bibitem[Ignatov et~al.(2017)Ignatov, Kobyshev, Timofte, Vanhoey, and
  Van~Gool]{ignatovDSLRQualityPhotosMobile2017a}
Andrey Ignatov, Nikolay Kobyshev, Radu Timofte, Kenneth Vanhoey, and Luc
  Van~Gool.
\newblock {DSLR}-quality photos on mobile devices with deep convolutional
  networks.
\newblock In \emph{Proceedings of the IEEE/CVF International Conference on
  Computer Vision}, pages 3277--3285, 2017.

\bibitem[Jiang et~al.(2021)Jiang, Dai, Wu, and
  Loy]{jiangFocalFrequencyLoss2021}
Liming Jiang, Bo Dai, Wayne Wu, and Chen~Change Loy.
\newblock Focal frequency loss for image reconstruction and synthesis.
\newblock In \emph{Proceedings of the IEEE/CVF International Conference on
  Computer Vision}, pages 13919--13929, 2021.

\bibitem[Johnson et~al.(2016)Johnson, Alahi, and
  Fei-Fei]{johnson2016perceptual}
Justin Johnson, Alexandre Alahi, and Li Fei-Fei.
\newblock Perceptual losses for real-time style transfer and super-resolution.
\newblock In \emph{European Conference on Computer Vision}, pages 694--711.
  Springer, 2016.

\bibitem[Kolouri et~al.(2017)Kolouri, Park, Thorpe, Slepcev, and
  Rohde]{kolouri2017optimal}
Soheil Kolouri, Se~Rim Park, Matthew Thorpe, Dejan Slepcev, and Gustavo~K
  Rohde.
\newblock Optimal mass transport: Signal processing and machine-learning
  applications.
\newblock \emph{IEEE Signal Processing Magazine}, 34\penalty0 (4):\penalty0
  43--59, 2017.

\bibitem[Krizhevsky et~al.(2012)Krizhevsky, Sutskever, and
  Hinton]{krizhevsky2012imagenet}
Alex Krizhevsky, Ilya Sutskever, and Geoffrey~E Hinton.
\newblock Imagenet classification with deep convolutional neural networks.
\newblock \emph{Advances in Neural Information Processing Systems}, 25, 2012.

\bibitem[Liang et~al.(2021)Liang, Cao, Sun, Zhang, Van~Gool, and
  Timofte]{liang2021swinir}
Jingyun Liang, Jiezhang Cao, Guolei Sun, Kai Zhang, Luc Van~Gool, and Radu
  Timofte.
\newblock Swinir: Image restoration using swin transformer.
\newblock In \emph{Proceedings of the IEEE/CVF Conference on Computer Vision
  and Pattern Recognition}, pages 1833--1844, 2021.

\bibitem[Liu et~al.(2021)Liu, Lin, Cao, Hu, Wei, Zhang, Lin, and
  Guo]{liu2021swin}
Ze Liu, Yutong Lin, Yue Cao, Han Hu, Yixuan Wei, Zheng Zhang, Stephen Lin, and
  Baining Guo.
\newblock Swin transformer: Hierarchical vision transformer using shifted
  windows.
\newblock In \emph{Proceedings of the IEEE/CVF International Conference on
  Computer Vision}, pages 10012--10022, 2021.

\bibitem[Mechrez et~al.(2018)Mechrez, Talmi, and
  Zelnik-Manor]{mechrezContextualLossImage2018}
Roey Mechrez, Itamar Talmi, and Lihi Zelnik-Manor.
\newblock The contextual loss for image transformation with non-aligned data.
\newblock In \emph{European Conference on Computer Vision}, pages 768--783.
  Springer, 2018.

\bibitem[Mei et~al.(2021)Mei, Fan, and Zhou]{mei2021image}
Yiqun Mei, Yuchen Fan, and Yuqian Zhou.
\newblock Image super-resolution with non-local sparse attention.
\newblock In \emph{Proceedings of the IEEE/CVF Conference on Computer Vision
  and Pattern Recognition}, pages 3517--3526, 2021.

\bibitem[Nguyen et~al.(2020)Nguyen, Ho, Pham, and
  Bui]{nguyenDISTRIBUTIONALSLICEDWASSERSTEINAPPLICATIONS2021}
Khai Nguyen, Nhat Ho, Tung Pham, and Hung Bui.
\newblock Distributional sliced-wasserstein and applications to generative
  modeling.
\newblock In \emph{International Conference on Learning Representations}, 2020.

\bibitem[Ni et~al.(2020)Ni, Yang, Wang, Ma, and Kwong]{ni2020towards}
Zhangkai Ni, Wenhan Yang, Shiqi Wang, Lin Ma, and Sam Kwong.
\newblock Towards unsupervised deep image enhancement with generative
  adversarial network.
\newblock \emph{IEEE Transactions on Image Processing}, 29:\penalty0
  9140--9151, 2020.

\bibitem[Ni et~al.(2022)Ni, Yang, Wang, Wang, Ma, and Kwong]{ni2022cycle}
Zhangkai Ni, Wenhan Yang, Hanli Wang, Shiqi Wang, Lin Ma, and Sam Kwong.
\newblock Cycle-interactive generative adversarial network for robust
  unsupervised low-light enhancement.
\newblock In \emph{Proceedings of the ACM International Conference on
  Multimedia}, pages 1484--1492, 2022.

\bibitem[Oppenheim and Lim(1981)]{phase2}
Alan~V Oppenheim and Jae~S Lim.
\newblock The importance of phase in signals.
\newblock \emph{Proceedings of the IEEE}, 69\penalty0 (5):\penalty0 529--541,
  1981.

\bibitem[Shen et~al.(2019)Shen, Wang, Lu, Shen, Ling, Xu, and Shao]{swapDeblur}
Ziyi Shen, Wenguan Wang, Xiankai Lu, Jianbing Shen, Haibin Ling, Tingfa Xu, and
  Ling Shao.
\newblock Human-aware motion deblurring.
\newblock In \emph{Proceedings of the IEEE/CVF International Conference on
  Computer Vision}, pages 5572--5581, 2019.

\bibitem[Simonyan and Zisserman(2015)]{Simonyan2014VeryDC}
K Simonyan and A Zisserman.
\newblock Very deep convolutional networks for large-scale image recognition.
\newblock In \emph{International Conference on Learning Representations}.
  Computational and Biological Learning Society, 2015.

\bibitem[Wang et~al.(2004)Wang, Bovik, Sheikh, and
  Simoncelli]{wangImageQualityAssessment2004}
Zhou Wang, Alan~C Bovik, Hamid~R Sheikh, and Eero~P Simoncelli.
\newblock Image quality assessment: From error visibility to structural
  similarity.
\newblock \emph{IEEE Transactions on Image Processing}, 13\penalty0
  (4):\penalty0 600--612, 2004.

\bibitem[Wei et~al.(2018)Wei, Wang, yang Wenhan, and Jiaying]{LOL}
Chen Wei, Wenjing Wang, yang Wenhan, and Liu Jiaying.
\newblock Deep retinex decomposition for low-light enhancement.
\newblock In \emph{British Machine Vision Conference}, 2018.

\bibitem[Yang and Soatto(2020)]{yang2020fda}
Yanchao Yang and Stefano Soatto.
\newblock {FDA}: Fourier domain adaptation for semantic segmentation.
\newblock In \emph{Proceedings of the IEEE/CVF Conference on Computer Vision
  and Pattern Recognition}, pages 4085--4095, 2020.

\bibitem[Zhang(2019)]{zhangMakingConvolutionalNetworks2019}
Richard Zhang.
\newblock Making convolutional networks shift-invariant again.
\newblock In \emph{International Conference on Machine Learning}, pages
  7324--7334, 2019.

\bibitem[Zhang et~al.(2018{\natexlab{a}})Zhang, Isola, Efros, Shechtman, and
  Wang]{kettunenELPIPSRobustPerceptual2019}
Richard Zhang, Phillip Isola, Alexei~A Efros, Eli Shechtman, and Oliver Wang.
\newblock The unreasonable effectiveness of deep features as a perceptual
  metric.
\newblock In \emph{Proceedings of the IEEE/CVF Conference on Computer Vision
  and Pattern Recognition}, pages 586--595, 2018{\natexlab{a}}.

\bibitem[Zhang et~al.(2018{\natexlab{b}})Zhang, Isola, Efros, Shechtman, and
  Wang]{zhangUnreasonableEffectivenessDeep2018}
Richard Zhang, Phillip Isola, Alexei~A Efros, Eli Shechtman, and Oliver Wang.
\newblock The unreasonable effectiveness of deep features as a perceptual
  metric.
\newblock In \emph{Proceedings of the IEEE/CVF Conference on Computer Vision
  and Pattern Recognition}, pages 586--595, 2018{\natexlab{b}}.

\bibitem[Zhang et~al.(2019)Zhang, Chen, Ng, and
  Koltun]{zhangZoomLearnLearn2019}
Xuaner Zhang, Qifeng Chen, Ren Ng, and Vladlen Koltun.
\newblock Zoom to learn, learn to zoom.
\newblock In \emph{Proceedings of the IEEE/CVF Conference on Computer Vision
  and Pattern Recognition}, pages 3762--3770, 2019.

\bibitem[Zhang et~al.(2018{\natexlab{c}})Zhang, Li, Li, Wang, Zhong, and
  Fu]{zhangImageSuperResolutionUsing2018}
Yulun Zhang, Kunpeng Li, Kai Li, Lichen Wang, Bineng Zhong, and Yun Fu.
\newblock Image super-resolution using very deep residual channel attention
  networks.
\newblock In \emph{European Conference on Computer Vision}, pages 286--301,
  2018{\natexlab{c}}.

\bibitem[Zhao et~al.(2016)Zhao, Gallo, Frosio, and Kautz]{zhao2016loss}
Hang Zhao, Orazio Gallo, Iuri Frosio, and Jan Kautz.
\newblock Loss functions for image restoration with neural networks.
\newblock \emph{IEEE Transactions on Computational Imaging}, 3\penalty0
  (1):\penalty0 47--57, 2016.

\bibitem[Zhou et~al.(2022)Zhou, Huang, Yan, Yu, Fu, Liu, Wei, and
  Zhao]{zhouSpatialFrequencyDomainInformation2022}
Man Zhou, Jie Huang, Keyu Yan, Hu Yu, Xueyang Fu, Aiping Liu, Xian Wei, and
  Feng Zhao.
\newblock Spatial-frequency domain information integration for pan-sharpening.
\newblock In \emph{European Conference on Computer Vision}, pages 274--291.
  Springer, 2022.

\end{thebibliography}
}


\end{document}